\documentclass[10pt,twocolumn,letterpaper]{article}

\usepackage{iccv}
\usepackage{times}
\usepackage{epsfig}
\usepackage{graphicx}
\usepackage{amsmath}
\usepackage{amssymb}
\usepackage{multirow}
\usepackage[numbers]{natbib}
\usepackage{paralist}
\usepackage{subcaption}

\newcommand*\rot{\rotatebox{90}}

\DeclareMathOperator*{\argmin}{arg\,min}

\usepackage[pagebackref=true,breaklinks=true,letterpaper=true,colorlinks,bookmarks=false]{hyperref}

\iccvfinalcopy 
\ificcvfinal\fi

\begin{document}

%%%%%%%%% TITLE
\title{Simultaneous Enhancement and Super-Resolution of Underwater \\ Imagery for Improved Visual Perception}

\author{Md Jahidul Islam, Peigen Luo and Junaed Sattar \\
{\tt\small \{islam034, luo00034, junaed\}@umn.edu} \\
{
\small Interactive Robotics and Vision Laboratory, Department of Computer Science and Engineering} \\ 
{
\small Minnesota Robotics Institute, University of Minnesota, Twin Cities, MN, USA }
}

\maketitle

\begin{abstract}
 In this paper, we introduce and tackle the simultaneous enhancement and super-resolution (\textbf{SESR}) problem for underwater robot vision and provide an efficient solution for near real-time applications. We present \textbf{Deep SESR}, a residual-in-residual network-based generative model that can learn to restore perceptual image qualities at $2\times$, $3\times$, or $4\times$ higher spatial resolution. We supervise its training by formulating a multi-modal objective function that addresses the chrominance-specific underwater color degradation, lack of image sharpness, and loss in high-level feature representation. It is also supervised to learn salient foreground regions in the image, which in turn guides the network to learn global contrast enhancement. We design an end-to-end training pipeline to jointly learn the saliency prediction and SESR on a shared hierarchical feature space for fast inference. Moreover, we present \textbf{UFO-120}, the first dataset to facilitate large-scale SESR learning; it contains over $1500$ training samples and a benchmark test set of $120$ samples. By thorough experimental evaluation on the UFO-120 and other standard datasets, we demonstrate that Deep SESR outperforms the existing solutions for underwater image enhancement and super-resolution. We also validate its generalization performance on several test cases that include underwater images with diverse spectral and spatial degradation levels, and also terrestrial images with unseen natural objects. Lastly, we analyze its computational feasibility for single-board deployments and demonstrate its operational benefits for visually-guided underwater robots. The model and dataset information will be available at:  \url{https://github.com/xahidbuffon/Deep-SESR}.
\end{abstract}

\vspace{-5mm}

\begin{figure}
    \centering
\includegraphics[width=\linewidth]{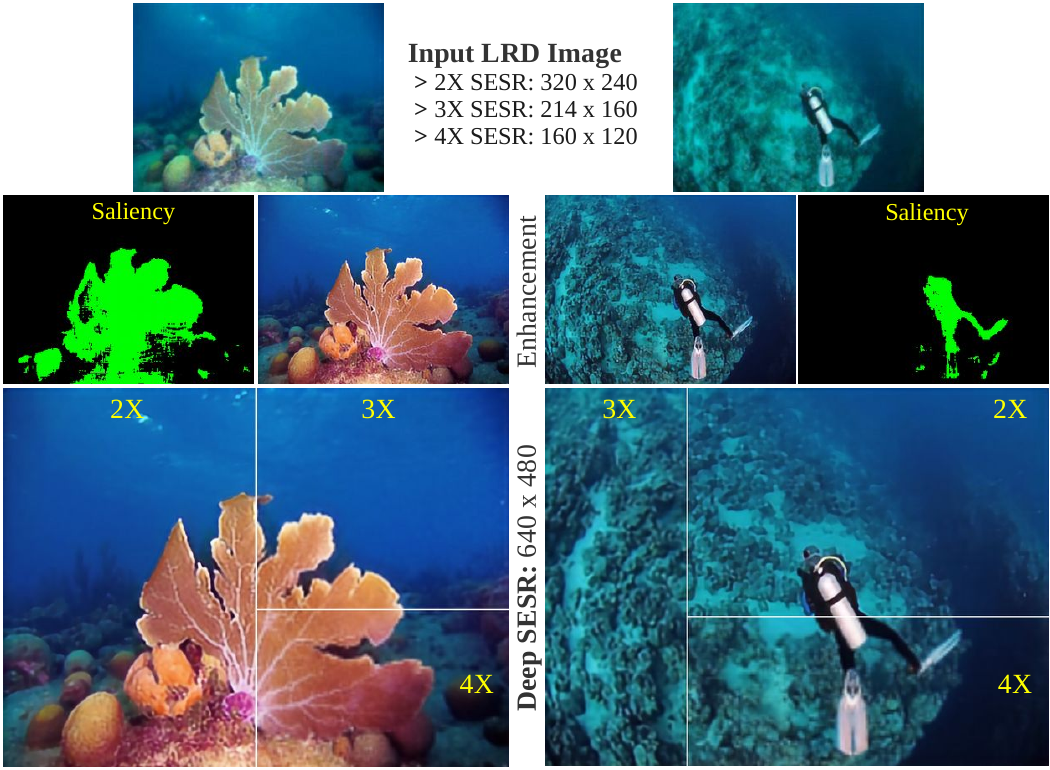}%
\vspace{-1mm}
 \caption{The proposed `Deep SESR' model offers perceptually enhanced HR image generation and saliency prediction by a single efficient inference. The enhanced images restore color, contrast, and sharpness at higher scales (up to $4\times$) to facilitate an improved visual perception, whereas the saliency map can be further exploited for attention modeling. \textit{All figures in this paper are best viewed digitally by zoom for colors and details. The dataset sources and credits for some online media resources (used in this work) are provided in Appendix I-II.}}%
 \label{fig:1}
    \vspace{-2mm}
\end{figure}

\section{Introduction}
Automatic generation of high resolution (HR) images from low resolution (LR) sensory measurements is a well-studied problem in the domains of computer vision and robotics due to its usefulness for detailed scene understanding and image synthesis~\cite{wang2018esrgan,zhang2018residual,islam2019underwater}. For visually-guided robots, in particular, this single image super-resolution (SISR) capability allows \emph{zooming-in} regions of interests (RoIs) for detailed perception, to eventually make navigational and other operational decisions.   
However, if the LR images suffer from noise and optical distortions, those get amplified by SISR, resulting in uninformative RoIs. Hence, restoring perceptual and statistical image qualities is essential for robust visual perception in noisy environments (\eg, underwater~\cite{bingham2010robotic,girdhar2014autonomous}). 
Although large bodies of literature on perceptual image enhancement and SISR offer solutions separately for both, a unified approach is more viable for computationally constrained real-time applications, which has not yet been explored in depth.

To this end, we introduce \textbf{simultaneous enhancement and super-resolution (SESR)}, and demonstrate its effectiveness for both underwater and terrestrial imagery. SESR is particularly useful in the underwater domain due to its unique optical properties ~\cite{akkaynak2018revised}, \eg, attenuation, refraction, and backscatter. These artifacts cause range-and-wavelength-dependent non-linear distortions that severely affect vision despite often using high-end cameras~\cite{islam2019fast}. Specifically, the captured images exhibit various levels of hue distortion, blurriness, low contrast, and color degradation based on the waterbody types, distances of light sources, etc. Some of these aspects can be modeled and estimated by physics-based solutions, particularly for dehazing~\cite{berman2018underwater}, color correction~\cite{bryson2016true}, water removal~\cite{akkaynak2019sea}, etc. However, these methods are often computationally too demanding for real-time robotic deployments. Besides, dense scene depth and optical waterbody measures are not always available in practical applications.

The learning-based approaches attempt to address the practicalities by approximating the underlying solution to the ill-posed problem of underwater image restoration with RGB data alone. Several existing models based on convolutional neural networks (CNNs)~\cite{liu2019underwater,wang2017deep} and generative adversarial networks (GANs)~\cite{islam2019fast,li2019fusion,fabbri2018enhancing} provide state-of-the-art (SOTA) performance for perceptual color enhancement, dehazing, deblurring, and contrast adjustment. Additionally, inspired by the success of deep residual networks for terrestrial SISR~\cite{zhang2018residual,ledig2017photo,hui2018fast}, several models have been proposed for underwater SISR in recent years~\cite{chen2019recovering,islam2019underwater}, which report exciting results with reasonable computational overhead. Contemporary research work~\cite{islam2019fast,islam2019underwater} further demonstrates that the perceptually enhanced underwater images provide significantly improved performance for widely-used object detection and human body-pose estimation tasks; moreover, detailed perception on salient image regions facilitates better scene understanding and attention modeling. However, as mentioned, separately processing visual data for these capabilities, even with the fastest available solutions, is not computationally feasible on single-board platforms.

In this paper, we present the first unified approach for SESR with an end-to-end trainable model. The proposed \textbf{Deep SESR} architecture incorporates dense residual-in-residual sub-networks to facilitate multi-scale hierarchical feature learning for SESR and saliency prediction. For supervision, we formulate a multi-modal objective function that evaluates the degree of chrominance-specific color degradation and loss in image sharpness, contrast, and high-level feature representation. As demonstrated in Fig.~\ref{fig:1}, it learns to restore perceptual image qualities at higher spatial scales (up to $4\times$); as a byproduct, it learns to identify salient foreground regions in the image. We also present the \textbf{UFO-120 dataset}, which contains over $1500$ annotated samples for large-scale SESR training, and a test set with an additional $120$ samples.

Furthermore, we evaluate the perceptual enhancement and super-resolution performance of Deep SESR on UFO-120 and several other standard datasets. The results suggest that it provides superior performance over SOTA methods on respective tasks, and achieves considerably better generalization performance on unseen natural images. It also achieves competitive performance on standard terrestrial datasets without additional training or tuning, which indicates that SESR methods can be potentially effective for terrestrial applications as well. Finally, we specify several design choices for Deep SESR, analyze their computational aspects, and discuss the usability benefits for its robotic deployments.

\section{Background}
%\subsection{Underwater Image Enhancement}
\textbf{Underwater image enhancement} is an active research problem that deals with correcting optical image distortions to recover true pixel intensities~\cite{akkaynak2019sea,bryson2016true}. Classical approaches use hand-crafted filters to improve local contrast and enforce color constancy. These approaches are inspired by the \textit{Retinex theory} of human visual perception~\cite{jobson1997multiscale,zhang2017underwater,fu2014retinex}, and mainly focus on restoring background illumination and lightness rendition. Another class of physics-based approaches uses an atmospheric dehazing model to estimate true \textit{transmission} and \textit{ambient} light in a scene~\cite{cho2018model,he2010single}.  Additional prior knowledge or statistical assumptions (\eg, haze-lines, dark channel prior~\cite{berman2018underwater}, etc.) are often utilized for global enhancements. Recent work by Akkaynak~\etal~\cite{akkaynak2018revised,akkaynak2019sea} introduces a revised image formation model that accounts for the unique characteristics of underwater light propagation; this contributes to a more accurate estimation of range-dependent attenuation and backscatter~\cite{roznere2019real}. 
%However, these methods require scene depth and optical water-body measures as prior. %, which are not always available in robotic applications. 
%Besides, these approaches tend to be computationally too demanding for real-time deployments. 
%Nevertheless, 

While accurate underwater image recovery remains a challenge, the learning-based approaches for \textit{perceptual enhancement} have made remarkable progress in recent years. Driven by large-scale supervised training~\cite{islam2019fast,yu2018underwater}, these approaches learn sequences of non-linear filters to approximate the underlying pixel-to-pixel mapping~\cite{isola2017image} between the \textit{distorted} and \textit{enhanced} image domains.  The contemporary deep CNN-based generative models provide SOTA performance in learning such image-to-image translation for both terrestrial~\cite{cheng2015deep,cai2016dehazenet} and underwater domains~\cite{islam2019fast,liu2019underwater}. Moreover, the GAN-based models attempt to improve generalization performance by employing a two-player min-max game~\cite{goodfellow2014generative}, where an adversarial \textit{discriminator} evaluates the \textit{generator}-enhanced images compared to ground truth samples. This forces the generator to learn realistic enhancement while evolving with the discriminator toward equilibrium. Several GAN-based underwater image enhancement models have reported impressive results from both paired~\cite{fabbri2018enhancing,li2018watergan} and unpaired training~\cite{islam2019fast}. However, they are prone to training instability, and hence require careful hyper-parameter choices, and intuitive loss function adaptation~\cite{arjovsky2017wasserstein,mao2017least} to ensure convergence.

%\subsection{Image Super-Resolution and Visual Attention Modeling}
\textbf{Single image super-resolution (SISR)} problem deals with automatically generating a sharp HR image from its LR measurements. Although SISR is relatively less studied in the underwater domain, a rich body of literature exists for terrestrial imagery~\cite{yang2019deep}. In particular, existing deep CNN-based models~\cite{dong2015image,ledig2017photo} and GAN-based models~\cite{sajjadi2017enhancenet,sonderby2016amortised} provide good solutions for SISR. Researchers have also exploited contemporary techniques~\cite{kim2016accurate,kim2016deeply,tong2017image} such as gradient clipping, dense skip connection, and sub-pixel convolution to improve SISR performance on standard datasets. Moreover, deep residual networks~\cite{ledig2017photo,hui2018fast} and residual-in-residual networks~\cite{wang2018esrgan,lim2017enhanced} are known to be very effective for learning SISR. Such networks employ skip connections to preserve the identity mapping within \textit{repeated blocks} of convolutional layers; this contributes to a stable training of very deep models. Zhang \etal~\cite{zhang2018residual} further demonstrated that dense skip connections within a residual block allow combining of hierarchical features from each layer, which substantially boosts the SISR performance. %Such residual dense blocks (RDBs) are used as building blocks of many state-of-the-art SISR models in a variety of applications.  

In recent years, similar ideas have been effectively applied for underwater imagery as well. For instance, Chen \etal~\cite{chen2019recovering} adopt residual-in-residual learning for underwater SISR, whereas Islam \etal~\cite{islam2019underwater} introduce a deep residual \textit{multiplier} model that can be dynamically configured for $2\times$, $4\times$, or $8\times$ SISR. Although these models report inspiring results, they do not account for underwater image distortions, and hence rely on a secondary network for enhancement. On the contrary, traditional approaches primarily focus on enhancing underwater image reconstruction quality by deblurring/denoising~\cite{chen2012model,quevedo2017underwater}, or descattering~\cite{lu2017underwater}. Hence, their applicability for end-to-end SESR is limited. 
%A few other research work attempt to improve fish recognition performance~\cite{sun2016fish} by learning better \textit{discriminating cues} at HR feature space. 
%Moreover, simultaneous feature learning for enhancement and super-resolution has not been explored in the literature.    

\textbf{Visual attention-based saliency prediction} refers to finding \textit{interesting} foreground regions in the image space~\cite{lu2016hierarchical,wang2017deep}. The classical \textit{stimulus-driven} approaches use features such as luminance, color, texture, and often depth information to quantify feature contrast in a scene. This feature contrast is subsequently exploited for spatial saliency computation. Automatic saliency prediction over a sequence of frames is also explored extensively~\cite{bazzani2016recurrent} because spatio-temporal features capture information about the motion and interaction among objects in a scene, which are important cues for attention modeling. Another genre of approaches deal with \textit{goal-driven} saliency prediction for visual question answering~\cite{yu2017multi},  \ie, finding the image regions that are relevant to a query.

In the underwater domain, however, existing research work mainly focuses on salient feature extraction for enhanced object detection performance~\cite{edgington2003automated,maldonado2016robotic,zhang2016underwater}. Hence, they do not provide a general solution for attention modeling that can facilitate faster visual search or better scene understanding. Nevertheless, finding salient RoIs in distorted underwater images and generating corresponding enhanced HR patches can be extremely useful for visually-guided robots. We attempt to contribute to these aspects in this paper. 

\section{Problem Formulation}
\subsection{Learning SESR}
SESR refers to the task of generating perceptually enhanced HR images from their LR and possibly distorted (LRD) input measurements. We formulate the problem as learning a pixel-to-pixel mapping from a source domain $X$ (of LRD images) to its target domain $Y$ (of enhanced HR images); we represent this mapping as a generative function $G: X \rightarrow Y$.
%As Figure~\ref{fig:prb} illustrates, w
We adopt an extended formulation by considering the task of learning SESR and saliency prediction on a shared feature space. Specifically, Deep SESR learns the generative function $G: X \rightarrow S, E, Y$; here, the additional outputs $S$ and $E$ denote the predicted saliency map, and enhanced image (in the same resolution as the input $X$), respectively. Additionally, it offers up to $4\times$ SESR for the final output $Y$. 

%\begin{figure}[h]
%    \centering
%    \vspace{-1mm}
%        \includegraphics[width=0.95\linewidth]{fig/prb_form.pdf}%
%        \vspace{-1mm}
%        \caption{The proposed Deep SESR model learns to generate three outputs from a LRD input: predicted saliency, enhanced image (same resolution), and SESR image (up to $4\times$ HR).}%
%        \vspace{-1mm}
%    \label{fig:prb}
%\end{figure}
%Hence, the objective is to learn   %As mentioned, this is an extremely ill-posed estimation problem without scene depth information, water-body types and underlying optical measurements. 

\begin{figure}[hb]
    \centering
    \begin{subfigure}{0.5\textwidth}
        \centering
        \includegraphics[width=\linewidth]{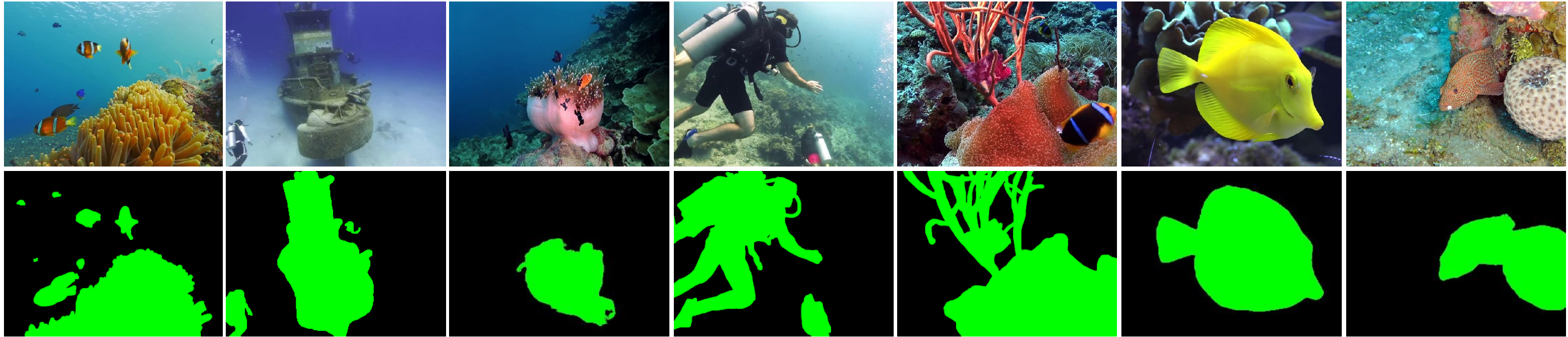}% 
        \vspace{-1mm}
        \caption{A few sample ground truth images and corresponding saliency maps are shown on the top, and bottom row, respectively.}
        \label{data_a}
    \end{subfigure}
    \vspace{1mm}
    
    \begin{subfigure}{0.5\textwidth} 
    \centering
        \includegraphics[width=\linewidth]{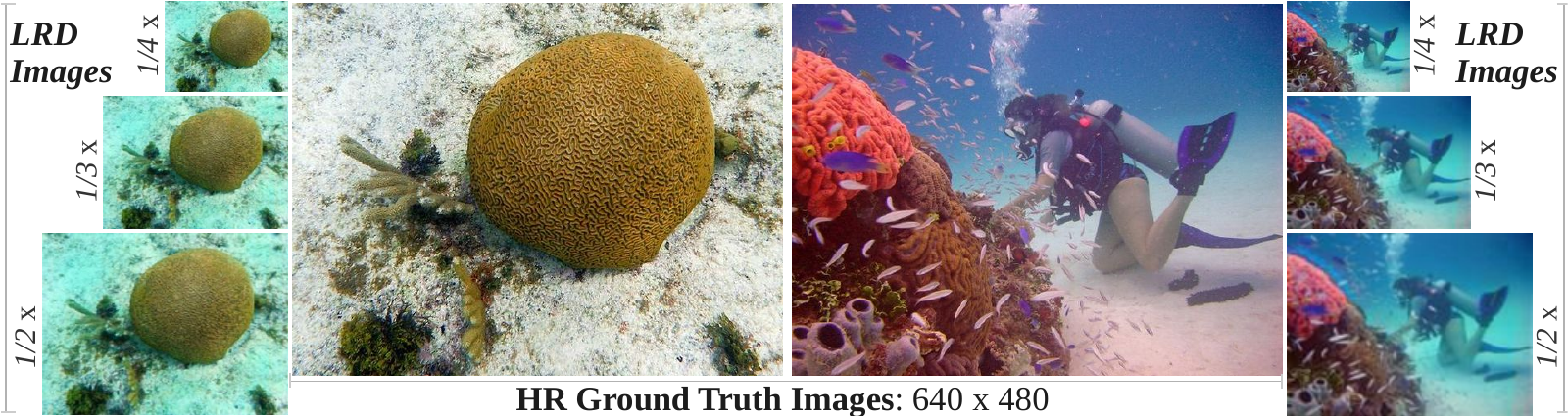} 
        \vspace{-1mm}
        \caption{Two particular instances are shown: the HR ground truth images are of size $640\times480$; their corresponding LR distorted (LRD) images are of size $320\times240$, $214\times160$, and $160\times120$. }
        \label{data_b}
    \end{subfigure}
    \vspace{-1mm}
    \caption{The UFO-120 dataset facilitates paired training of $2\times$, $3\times$, and $4\times$ SESR models; it also contains salient pixel annotations for all training samples. The combined data is used for the supervised training of Deep SESR model.}
    \vspace{-2mm}
    \label{fig:data}
\end{figure}

\subsection{Data Preparation: The UFO-120 Dataset}\label{sec:data_prep}
We utilize several existing underwater image enhancement and super-resolution datasets to supervise the SESR learning. We follow standard procedures~\cite{islam2019fast,li2019feedback,fabbri2018enhancing} for optical/spatial image degradation, and use human-labeled saliency maps to create paired data of the form ($\{X\}$, $\{S, E, Y\}$); further details on the existing datasets are provided in Section~\ref{implement}.

In addition, we contribute over $1500$ samples for training (and another $120$ for testing) in the UFO-120 dataset. It contains images collected from oceanic explorations in multiple locations having different water types, as seen in Fig.~\ref{data_a}. The salient foreground pixels of each image are annotated by human participants. Moreover, we adopt a widely used style-transfer technique~\cite{islam2019fast,fabbri2018enhancing} to generate their respective distorted images. Subsequently, we generate the LRD samples by Gaussian blurring ({\tt GB}) and bicubic down-sampling ({\tt BD}); based on their relative order, we group the data into three sets:     
\begin{itemize}
    \item \textbf{Set-U:} {\tt GB} is followed by {\tt BD}.
    \item \textbf{Set-F:} the order is interchanged with a $0.5$ probability. 
    \item \textbf{Set-O:} {\tt BD} is followed by {\tt GB}.
\end{itemize}

We use a $7\times7$ kernel and a noise level of $20\%$ for {\tt GB}. Additionally, as Fig.~\ref{data_b} illustrates, we use $2\times$, $3\times$, and $4\times$ {\tt BD} to generate the LRD samples. Hence, there are nine available training combinations for SESR. The UFO-120 dataset can also be used for training underwater SISR ($E$$\rightarrow$$Y$), image enhancement ($X$$\rightarrow$$E$), or saliency prediction ($E$$\rightarrow$$S$) models. %The data and relevant information will be made public for academic research purposes after the blind-review process. 

\begin{figure}[t]
    \centering
    \begin{subfigure}{0.5\textwidth}
    \centering
        \includegraphics[width=0.9\linewidth]{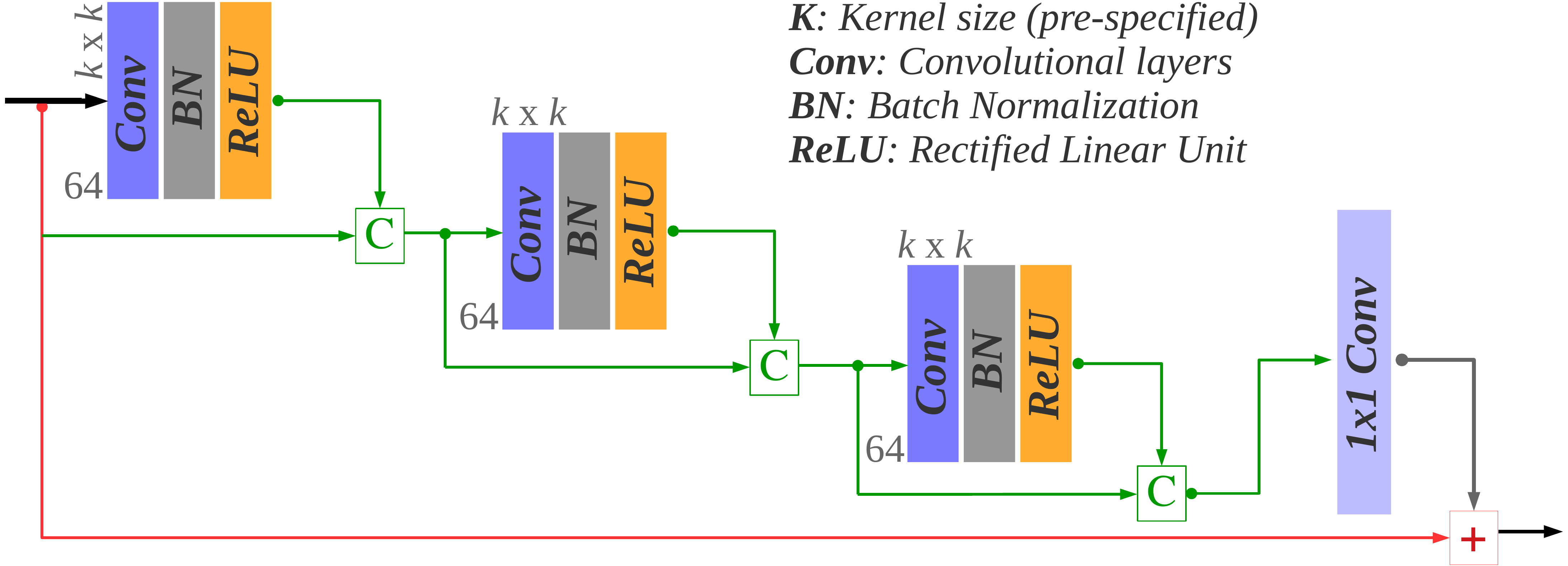}%
        \vspace{-1mm}
        \caption{A residual dense block (RDB)~\cite{zhang2018residual}.}
        \label{fig:model_rdb}
    \end{subfigure}
    \vspace{2mm}
    
    \begin{subfigure}{0.5\textwidth}
    \centering
        \includegraphics[width=\linewidth]{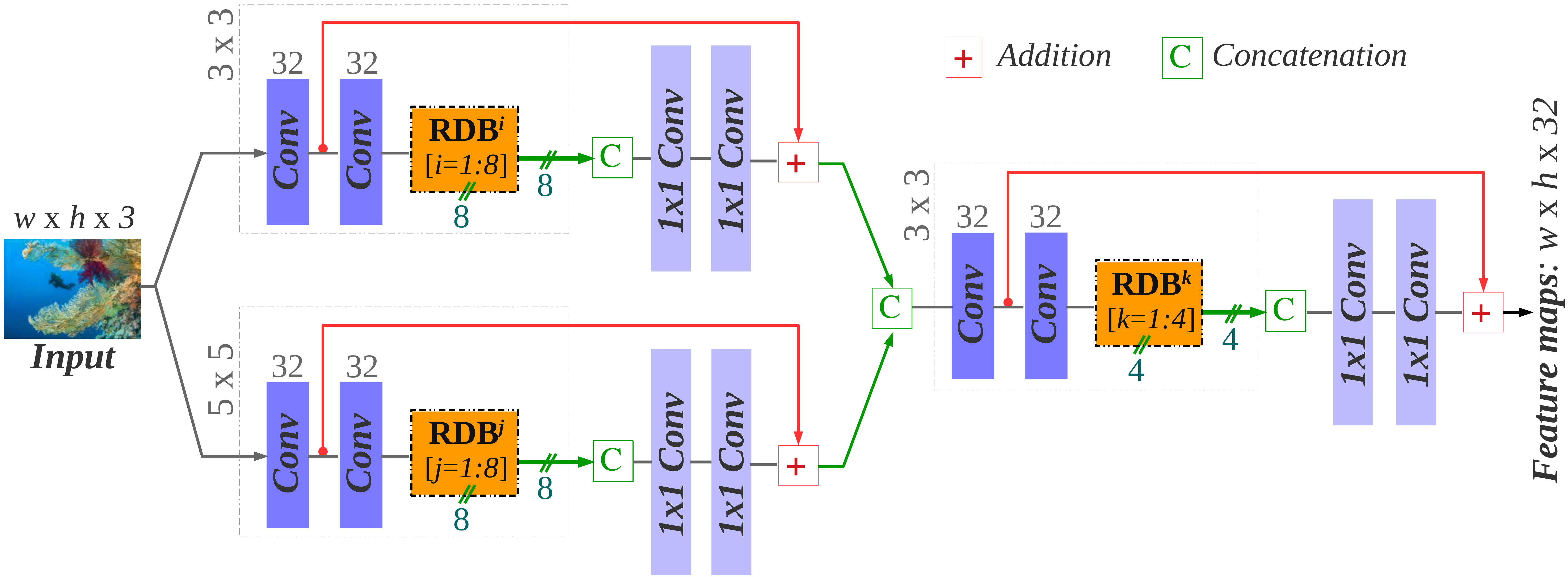}%
         \vspace{-1mm}
        \caption{The feature extraction network (FENet).}
        \label{fig:model_fnet}
    \end{subfigure}
    \vspace{2mm}
    
    \begin{subfigure}{0.5\textwidth}
    \centering
    \includegraphics[width=\linewidth]{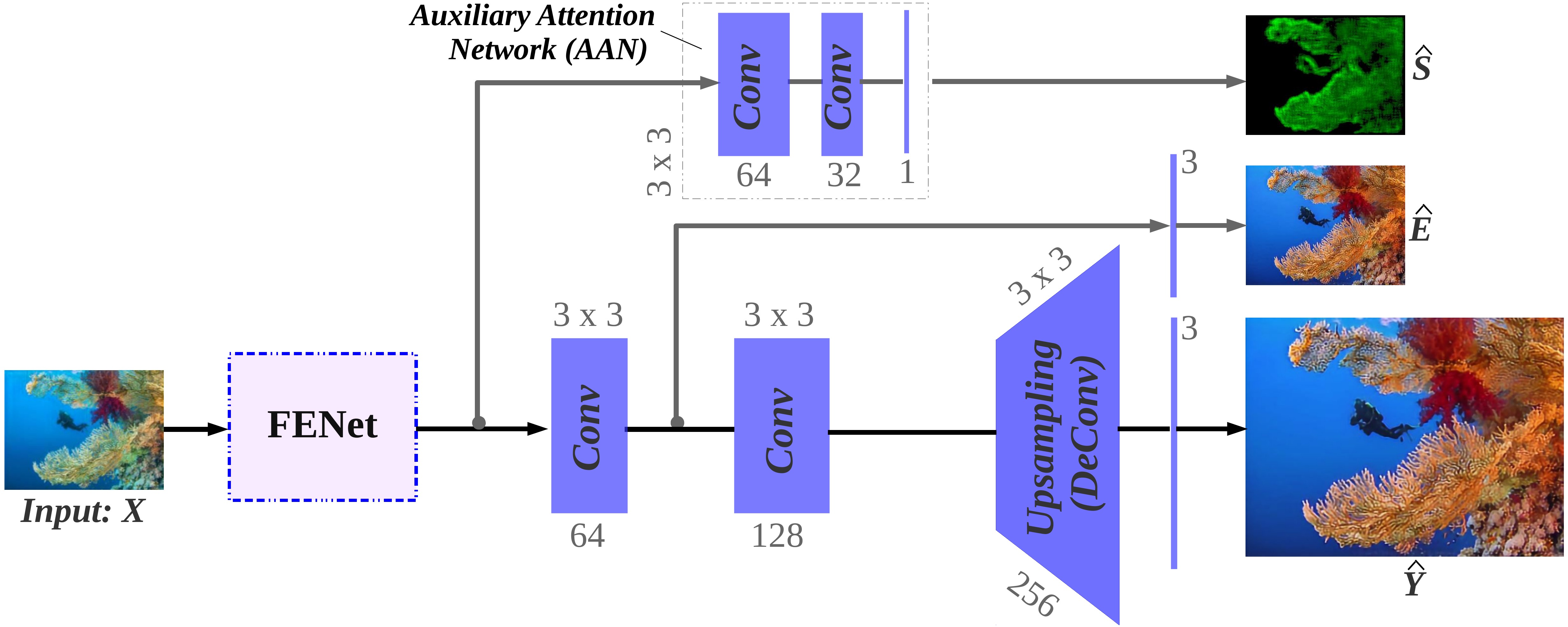}%
    \vspace{-1mm}
    \caption{The end-to-end architecture is shown. FENet-extracted feature maps are propagated along two branches: \textit{i)} to AAN for learning saliency, and \textit{ii)} to an intermediate convolutional layer for learning enhancement. Another convolutional layer and subsequent upsampling layers learn SESR along the main branch.}
    \label{fig:deep_sesr}
    \end{subfigure}
    \vspace{-1mm}
    \caption{Network architecture and detailed parameter specification of the proposed Deep SESR model.}
    \vspace{-3mm}
    \label{fig:model}
\end{figure}

\section{Deep SESR Model}
\subsection{Network Architecture}
As shown in Figure~\ref{fig:model}, the major components of our Deep SESR model are: residual dense blocks (RDBs), a feature extraction network (FENet), and an auxiliary attention network (AAN). These components are tied to an end-to-end architecture for the combined SESR learning.

\textbf{Residual Dense Blocks (RDBs)} consist of three sets of convolutional ({\tt conv}) layers, each followed by Batch Normalization ({\tt BN})~\cite{ioffe2015batch} and {\tt ReLU} non-linearity~\cite{nair2010rectified}. As Figure~\ref{fig:model_rdb} illustrates, the input and output of each layer is concatenated to subsequent layers. This architecture is inspired by Zhang \etal~\cite{zhang2018residual} who demonstrated that such dense skip connections facilitate an improved hierarchical feature learning. Each {\tt conv} layer learns $64$ filters of a given kernel size; their outputs are then fused by a $1\times1$ {\tt conv} layer for local residual learning.

\textbf{Feature Extraction Network (FENet)} uses RDBs as building blocks to incorporate two-stage residual-in-residual learning. As shown in Figure~\ref{fig:model_fnet}, on the first stage, two parallel branches use eight RDB blocks each to separately learn $3\times3$ and $5\times5$ filters in input image space; these filters are then concatenated and passed to a common branch for the second stage of learning. Four RDB blocks with $3\times3$ filters are used in the later stage which eventually generates $32$ feature maps. % as the final output. 
Our motive for such design is to have the capacity to learn locally dense informative features while still maintaining a globally shallow architecture to ensure fast feature extraction.
%during inference.               

\textbf{Auxiliary Attention Network (AAN)} learns to model visual attention in the FENet-extracted feature space. As shown in Figure~\ref{fig:deep_sesr}, two sequential {\tt conv} layers learn to generate a single channel output that represents saliency (probabilities) for each pixel. We show the predicted saliency map as \textit{green} intensity values; the black pixels represent background regions.

\textbf{The Deep SESR learning} is guided along the primary branch by a series of {\tt conv} and {\tt deconv} (de-convolutional) layers. As Figure~\ref{fig:deep_sesr} demonstrates, the enhanced image (LR), and the SESR image (HR) are generated by separate output layers at different stages in the network. The enhanced image is generated from the {\tt conv} layer that immediately follows FENet; it is supervised to learn enhancement by dedicated loss functions applied at the shallow output layer. The \textit{enhanced} features are also propagated to another {\tt conv} layer, followed by {\tt deconv} layers for upsampling. The final SESR output is generated from upsampled features based on the given scale: $2\times$, $3\times$, or $4\times$. %The loss functions applied at the final layer and
Other model parameters, \eg, the number of filters, kernel sizes, etc., are annotated in Figure~\ref{fig:model}.

\subsection{Loss Function Formulation}\label{loss_fun}
The end-to-end training of Deep SESR is supervised by seven loss components that address various aspects of learning the function $G: X\rightarrow S, E, Y$. By denoting $\hat{S},\hat{E},\hat{Y} = G(X)$ as the generated output, we formulate the loss terms as follows:

\textit{1)} \textbf{Information Loss} for saliency prediction is measured by a standard cross-entropy function~\cite{lu2016hierarchical,wang2017deep}. 
It quantifies the dissimilarity in pixel intensity distributions between the generated saliency map ($\hat{S}$) and its ground truth ($S$). For a total of $N_p$ pixels in $\hat{S}$, it is calculated as   
\begin{equation}
    \mathcal{L}_{Saliency}^{AAN} = \frac{1}{N_p}  \sum_{p=1}^{N_p}\big[ -S_p \log \hat{S}_p - (1-S_p) \log (1-\hat{S}_p) \big].
\end{equation}

\textit{2)} \textbf{Contrast loss (LR)} evaluates the hue and luminance recovery in the enhanced images. The dominating green/blue hue in distorted underwater images often causes low-contrast and globally dim foreground pixels. We quantify this loss of relative strength (\ie, intensity) in foreground pixels in RGB space by utilizing a differentiable function: Contrast Measurement Index (CMI)~\cite{shaus2017potential,trivedi2013novel}. The CMI measures the average intensity of foreground pixels ($F_I$) relative to the background ($B_I$) for an image $I$, as 
\[ CMI(I) = \frac{(F_I - B_I)}{(F_I + B_I)} \text{ } \propto (F_I - B_I).\] 
We exploit the saliency map $S$ (or $\hat{S})$ to find the foreground pixels in $E$ (or $\hat{E})$, as $F_E = E\odot S$ and $F_{\hat{E}} = \hat{E}\odot \hat{S}$; here, $\odot$ denotes element-wise multiplication. Subsequently, we compute the contrast loss as 
\begin{equation} 
    \mathcal{L}_{Contrast}^{LR} = \big|\big| CMI(E)-CMI(\hat{E}) \big|\big|_2.
\end{equation}

\begin{figure}[hb]
    \centering
    \begin{subfigure}{0.3\textwidth}
    \centering
        \includegraphics[width=\linewidth]{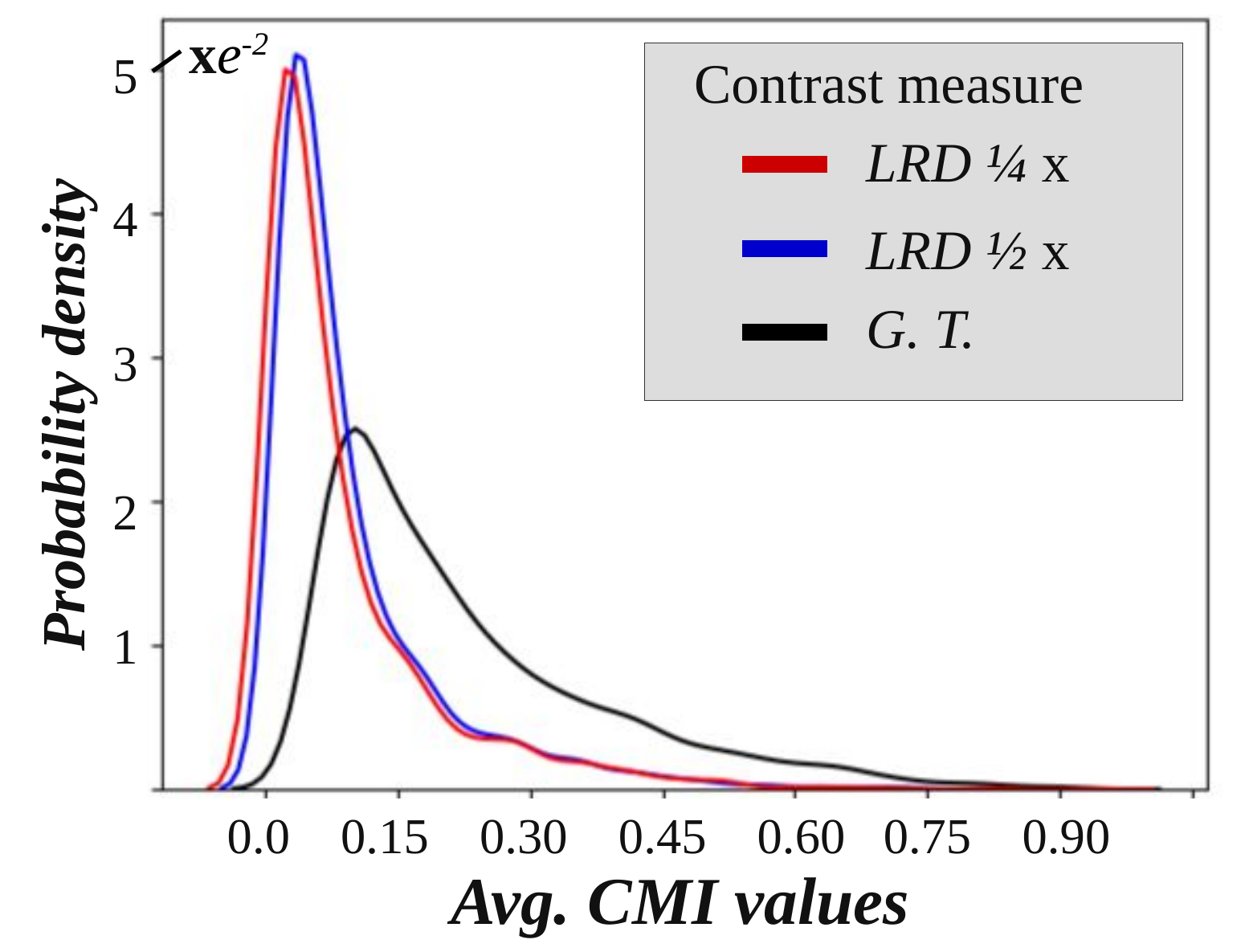}%
        \vspace{-1mm}
        \caption{Contrast measure: $CMI(I)$.}%
        \label{fig:con_a}
    \end{subfigure}
    \vspace{3mm}
    
    \begin{subfigure}{0.3\textwidth}
        \centering
        \includegraphics[width=\linewidth]{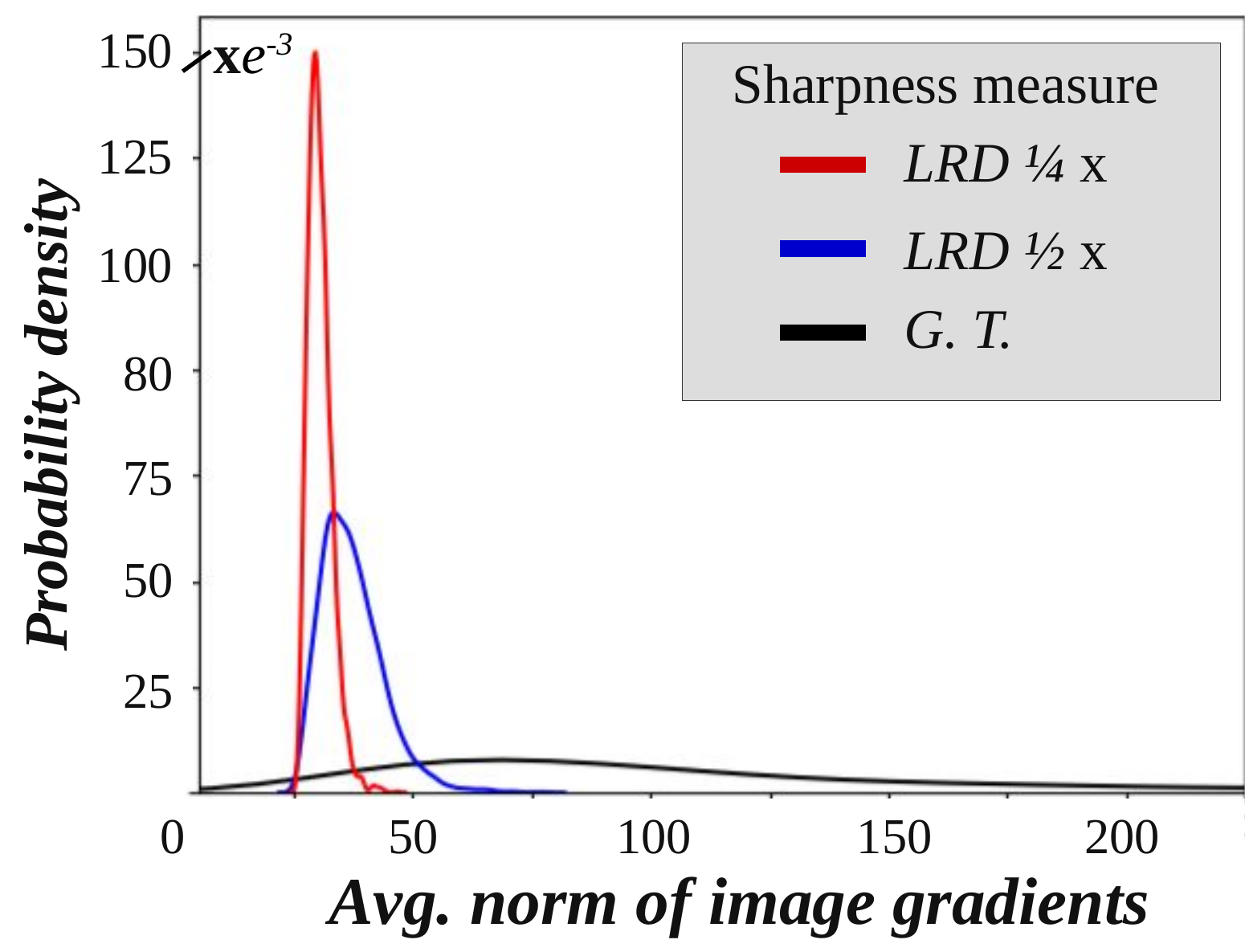}%
        \vspace{-1mm}
        \caption{Sharpness measure: $|\nabla I|$.}%
        \label{fig:sharp_b}
    \end{subfigure}
    \vspace{3mm}
    
        \begin{subfigure}{0.5\textwidth}
        \centering
        \includegraphics[width=\linewidth]{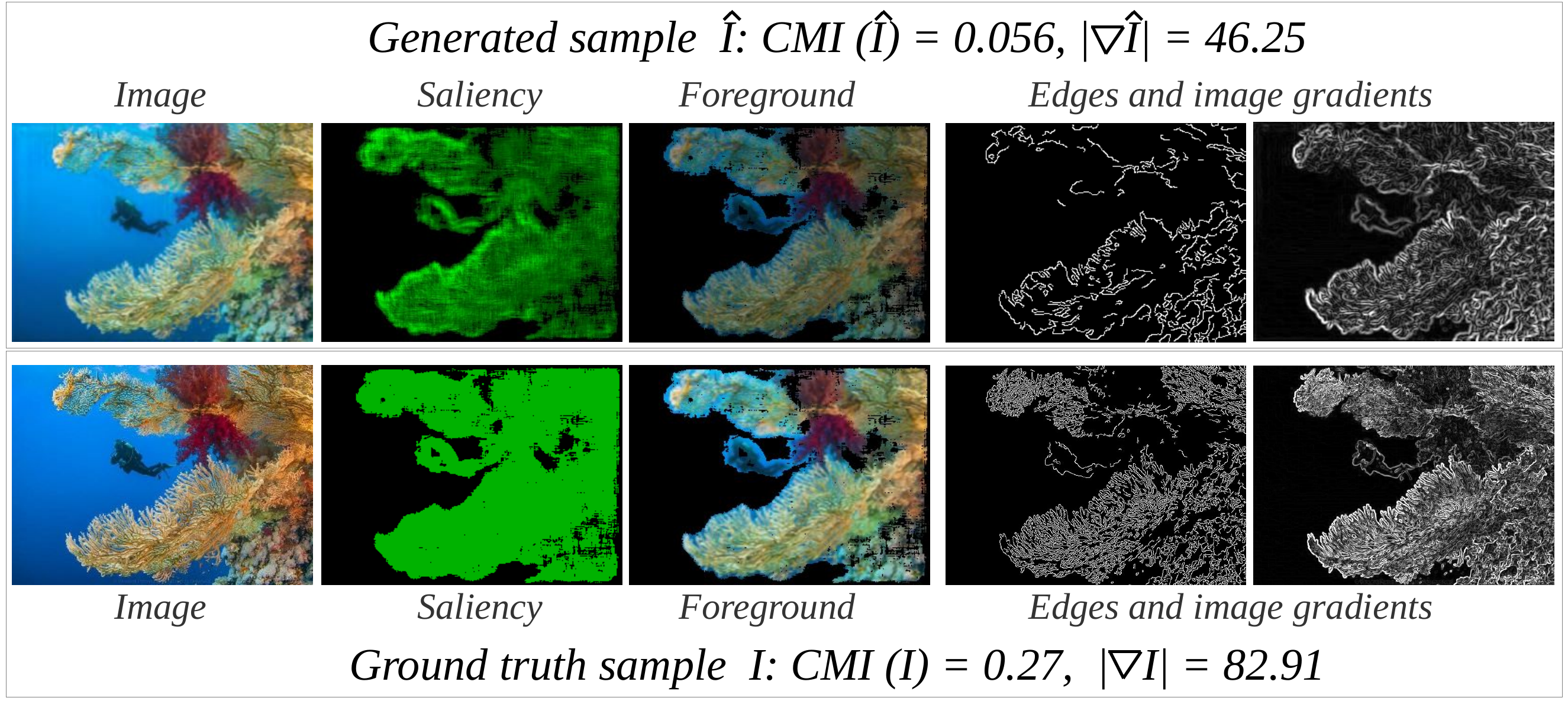}%
        \vspace{-1mm}
        \caption{Image contrast and sharpness properties of a particular sample compared to its ground truth measurement.}
        \label{fig:cons_sharp_exp}
    \end{subfigure}%
    \caption{The lack of contrast and sharpness in LRD samples of UFO-120 dataset (compared to their ground truth) are shown in (a) and (b); as seen, distributions for LRD samples are densely skewed to lower values, whereas the ground truth distributions span considerably higher values. A qualitative interpretation of this numeric disparity is illustrated in (c).}
    \vspace{-2mm}
    \label{fig:loss}
\end{figure}

An immediate consequence of using $\mathcal{L}_{Contrast}^{LR}$ is that AAN can directly influence learning enhancement despite being on a separate branch. Such coupling also provides better training stability (otherwise AAN tends to converge too early and starts over-fitting). Moreover, in Figure~\ref{fig:con_a}, we show the distributions of CMI for training samples of the UFO-120 dataset, which suggests that the distorted samples' CMI scores are skewed to much lower values compared to the ground truth. Hence, $\mathcal{L}_{Contrast}^{LR}$ forces the CMI distribution to shift toward higher values for learning contrast enhancement.

\textit{3)} \textbf{Color Loss (LR/HR)} evaluates global similarity of the enhanced ($\hat{E}$) and SESR output ($\hat{Y}$) with respective ground truth measurements in RGB space. The standard $\mathcal{L}_{2}$ loss terms are: $	\mathcal{L}_{2}^{LR} = \big|\big|E-\hat{E}\big|\big|_2, \text{ and  } \mathcal{L}_{2}^{HR} = \big|\big|Y-\hat{Y}\big|\big|_2$. Additionally, we formulate two \textit{perceptual loss} functions that are particularly designed for learning underwater image enhancement and super-resolution. First, we utilize two wavelength-dependent chrominance terms: $C_{rg}=(r- g)$, and $C_{yb}=\frac{1}{2}(r+g)-b$, which are core elements of the Underwater Image Colorfulness Measure (UICM)~\cite{panetta2016human,liu2019real}. By denoting $\Delta r$, $\Delta g$, and $\Delta b$, as the per-channel numeric differences between $\hat{E}$ and $E$, we formulate the loss as:  
\begin{equation}
	\mathcal{L}_{P}^{LR} = \big|\big| 4{ (\Delta r - \Delta g})^2 + ({\Delta r + \Delta  g - 2\Delta b})^2  \big|\big|_2.
\end{equation}
On the other hand, being inspired by~\cite{compuphase,islam2019underwater}, we evaluate the perceptual similarity at HR as   
\begin{equation}
	\mathcal{L}_{P}^{HR} = \big|\big| \frac{(512+\bar{R})}{256} \Delta R^2 + 4 \Delta G^2+ \frac{(767-\bar{R})}{256}\Delta B^2  \big|\big|_2.
\end{equation}
Here, $\bar{R} = (R_{Y} + R_{\hat{Y}})/2$, whereas $\Delta R$, $\Delta G$, and $\Delta B$ are the per-channel disparities between $\hat{Y}$ and $Y$. Finally, we adopt the color loss terms for enhancement and SESR as   
\begin{align}
    \mathcal{L}_{Color}^{LR} &= 0.25 \text{ } \mathcal{L}_{P}^{LR} + 0.75 \text{ } \mathcal{L}_{2}^{LR}, \text{ and } \\
    \mathcal{L}_{Color}^{HR} &= 0.25 \text{ } \mathcal{L}_{P}^{HR} + 0.75 \text{ } \mathcal{L}_{2}^{HR},\text{ respectively.}
\end{align}

\textit{4)} \textbf{Content loss (LR/HR)} forces the generator to restore a similar \textit{feature content} as the ground truth in terms of high-level representation. Such feature preservation has been found to be very effective for image enhancement, style transfer, and SISR problems~\cite{ignatov2017dslr,islam2019underwater}; as suggested in~\cite{johnson2016perceptual}, we define the image content function $\Phi_{VGG}(\cdot)$ as high-level features extracted by the last {\tt conv} layer of a pre-trained VGG-19 network. Then, we formulate the content loss for enhancement and SESR as
\begin{align}
    \mathcal{L}_{Content}^{LR} &= \big|\big|\Phi_{VGG} (E) - \Phi_{VGG} (\hat{E}) \big|\big|_2, \text{ and } \\
    \mathcal{L}_{Content}^{HR} &= \big|\big|\Phi_{VGG} (Y) - \Phi_{VGG} (\hat{Y}) \big|\big|_2,\text{ respectively.}
\end{align}

\textit{5)} \textbf{Sharpness loss (HR)} measures the blurriness recovery in SESR output by exploiting local image gradients.  The literature offers several solutions for evaluating image sharpness based on norm/histogram of gradients or frequency-domain analysis. In particular, the notions of Just Noticable Blur (JNB)~\cite{ferzli2009no} and Perceptual Sharpness Index (PSI)~\cite{feichtenhofer2013perceptual} are widely used; they apply non-linear transformation and thresholding on local contrast or gradient-based features to quantify perceived blurriness based on the characteristics of human visual system.   
However, we found better results and numeric stability by using the norm of image gradients directly; specifically, we use the standard $3\times3$ Sobel operator~\cite{gao2010improved} for computing spatial gradient $\nabla I = \sqrt{I_x^2 + I_y^2}$ for an image $I$. Subsequently, we formulate the sharpness loss for SESR as     
\begin{equation} 
    \mathcal{L}_{Sharpness}^{HR} = \big|\big| \text{ } |\nabla Y|^2 - |\nabla \hat{Y}|^2 \text{ } \big|\big|_1.
\end{equation}

In Figure~\ref{fig:sharp_b}, we present a statistical validity of $\mathcal{L}_{sharpness}^{LR}$ as a loss component; also, edge gradient features for a particular sample are provided in Figure~\ref{fig:cons_sharp_exp}. As shown, numeric disparities for the norm of gradients between distorted images and their HR ground truth are significant, which we quantify by $\mathcal{L}_{sharpness}^{LR}$ to encourage sharper image generation.

\subsection{End-to-end Training Objective}
We use a linear combination of the above-mentioned loss components to formulate the unified objective function as   
\begin{align} \label{eq:final}
\centering
	G^* = \argmin\limits_{G} \big\{ 
	 \lambda_s^{AAN} \mathcal{L}_{Saliency}^{AAN} + \mathcal{L}_{SESR}^{LR} + \mathcal{L}_{SESR}^{HR}  \big\}\text{;}  
\end{align}
where $\mathcal{L}_{SESR}^{LR}$ and $\mathcal{L}_{SESR}^{HR}$ are expressed by   
\begin{align*} %\label{eq:pix2pix_final}
\centering
	\mathcal{L}_{SESR}^{LR} &= 
	 \lambda_c^{LR} \mathcal{L}_{Color}^{LR}+\lambda_f^{LR}  \mathcal{L}_{Content}^{LR}+\lambda_t^{LR} \mathcal{L}_{Contrast}^{LR},\text{ and} \\
	\mathcal{L}_{SESR}^{HR} &= 
	 \lambda_c^{HR}  \mathcal{L}_{Color}^{HR}+\lambda_f^{HR} \mathcal{L}_{Content}^{HR}+\lambda_g^{HR} \mathcal{L}_{Sharpness}^{HR}.
\end{align*}
Here, $\lambda_{\odot}^{\Box}$ symbols are scaling factors that represent the contributions of respective loss components; their values are empirically tuned as hyper-parameters.

\begin{figure*}
\vspace{-1mm}
    \centering
        \includegraphics[width=\linewidth]{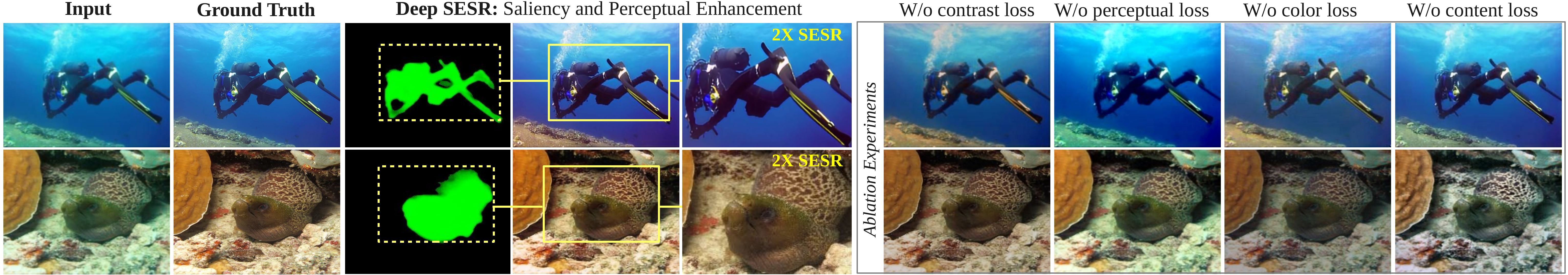}%
        \vspace{-1mm}
        \caption{Each row demonstrates perceptual enhancement and saliency prediction by Deep SESR on respective LRD input images; the corresponding results of an ablation experiment shows contributions of various loss-terms in the learning.}%
    \label{fig:qual_en}
    \vspace{2mm}
\end{figure*}

\begin{table*}
\centering
\caption{Quantitative performance comparison for enhancement: scores are shown as $\textit{mean} \pm \sqrt{\textit{variance}}$; the first and second best scores (in each row) are colored red, and blue, respectively.}
%\footnotesize
\scriptsize
\vspace{-1mm}
\begin{tabular}{c|l||c|c|c|c|c|c|c|c|c}
  \cline{1-11}
  & \textbf{Dataset} & RGHS & UCM & MS-Fusion & MS-Retinex & Water-Net & UGAN & Fusion-GAN & FUnIE-GAN & \textbf{Deep SESR}\\ \hline \hline
  \parbox[t]{1mm}{\multirow{3}{*}{\rot{\textit{PSNR}}}} & UFO-120 & $20.05 \pm 3.1$ & $20.99 \pm 2.2$ & $21.32 \pm 3.3$ & $21.69 \pm 3.6$ & $22.46 \pm 1.9$ & $23.45 \pm 3.1$ & $24.07 \pm 2.1$ & {\color{blue}$25.15 \pm 2.3$} & {\color{red}$27.15 \pm 3.2$} \\ %\hline
  & EUVP & $20.12 \pm 2.9$ & $20.55 \pm 1.8$  & $19.85 \pm 2.4$ & $21.27 \pm 3.1$ & $20.14 \pm 2.3$ & $23.67 \pm 1.5$ & $23.77 \pm 2.4$ & {\color{red}$26.78 \pm 1.1$} & {\color{blue}$25.25 \pm 2.1$} \\ %\hline
  & UImNet & $19.98 \pm 1.8$  & $20.48 \pm 2.2$ & $19.59 \pm 3.2$ & $22.63 \pm 2.5$ & $21.02 \pm 1.6$ & $23.88 \pm 2.1$ & $23.12 \pm 1.9$ & {\color{blue}$24.68 \pm 2.4$} & {\color{red}$25.52 \pm 2.7$} \\ 
  \hline
  \parbox[t]{1mm}{\multirow{3}{*}{\rot{\textit{SSIM}}}} & UFO-120 & $0.75 \pm 0.06$ & $0.78 \pm 0.07$ & $0.79 \pm 0.09$ & $0.75 \pm 0.10$ & $0.79 \pm 0.05$ & $0.80 \pm 0.08$ & {\color{blue}$0.82 \pm 0.07$} & {\color{blue}$0.82 \pm 0.08$} & {\color{red}$0.84 \pm 0.03$} \\ %\hline
  & EUVP &  $0.69 \pm 0.11$ & $0.73 \pm 0.14$  & $0.70 \pm 0.05$ & $0.69 \pm 0.15$ & $0.68 \pm 0.18$ & $0.67 \pm 0.11$ & $0.68 \pm 0.05$ & {\color{red}$0.86 \pm 0.05$} & {\color{blue}$0.75 \pm 0.07$} \\ %\hline
  & UImNet & $0.61 \pm 0.08$  & $0.67 \pm 0.06$ & $0.64 \pm 0.11$ & $0.74 \pm 0.04$ & $0.71 \pm 0.07$ & {\color{blue}$0.79 \pm 0.08$} & $0.75 \pm 0.07$ & $0.77 \pm 0.06$ & {\color{red}$0.81 \pm 0.05$} \\ 
  \hline
  \parbox[t]{1mm}{\multirow{3}{*}{\rot{\textit{UIQM}}}} & UFO-120 & $2.36 \pm 0.33$ & $2.41 \pm 0.53$ & $2.76 \pm 0.45$ & $2.69 \pm 0.59$ & $2.83 \pm 0.48$ & $3.04 \pm 0.28$ & $2.98 \pm 0.28$ & {\color{blue}$3.09 \pm 0.51$} & {\color{red}$3.13 \pm 0.45$} \\ 
  & EUVP &  $2.45 \pm 0.46$ &  $2.48 \pm 0.77$ & $2.51 \pm 0.36$ & $2.48 \pm 0.09$ & $2.55 \pm 0.06$ & $2.70 \pm 0.31$ & $2.58 \pm 0.07$ & {\color{blue}$2.95 \pm 0.38$} & {\color{red}$2.98 \pm 0.28$} \\ 
  & UImNet & $2.32 \pm 0.48$ & $2.38 \pm 0.42$ & $2.79 \pm 0.55$ & $2.84 \pm 0.37$ & $2.92 \pm 0.35$ & {\color{red}$3.32 \pm 0.55$} & $3.19 \pm 0.27$ & $3.23 \pm 0.32$ & {\color{blue}$3.26 \pm 0.36$} \\ 
  %EUVP$_{up}$ &  $2.27 \pm 0.66$ &  & & & & & & & \\ 
  %& U45$^\dagger$ & $2.65 \pm 0.75$  & $2.67 \pm 0.70$ & $2.88 \pm 0.39$ & $2.65 \pm 0.70$  &  $2.96 \pm 0.71$ & {\color{blue}$\mathbf{3.38 \pm 0.17}$} & {\color{blue}${3.27 \pm 0.26}$} &  $3.08 \pm 0.46$ & $2.98 \pm 0.54$  \\
 \hline
\end{tabular}
\label{tab:quan_se}
\vspace{1mm}
\end{table*}  

\begin{figure*}
    \centering
        \includegraphics[width=\linewidth]{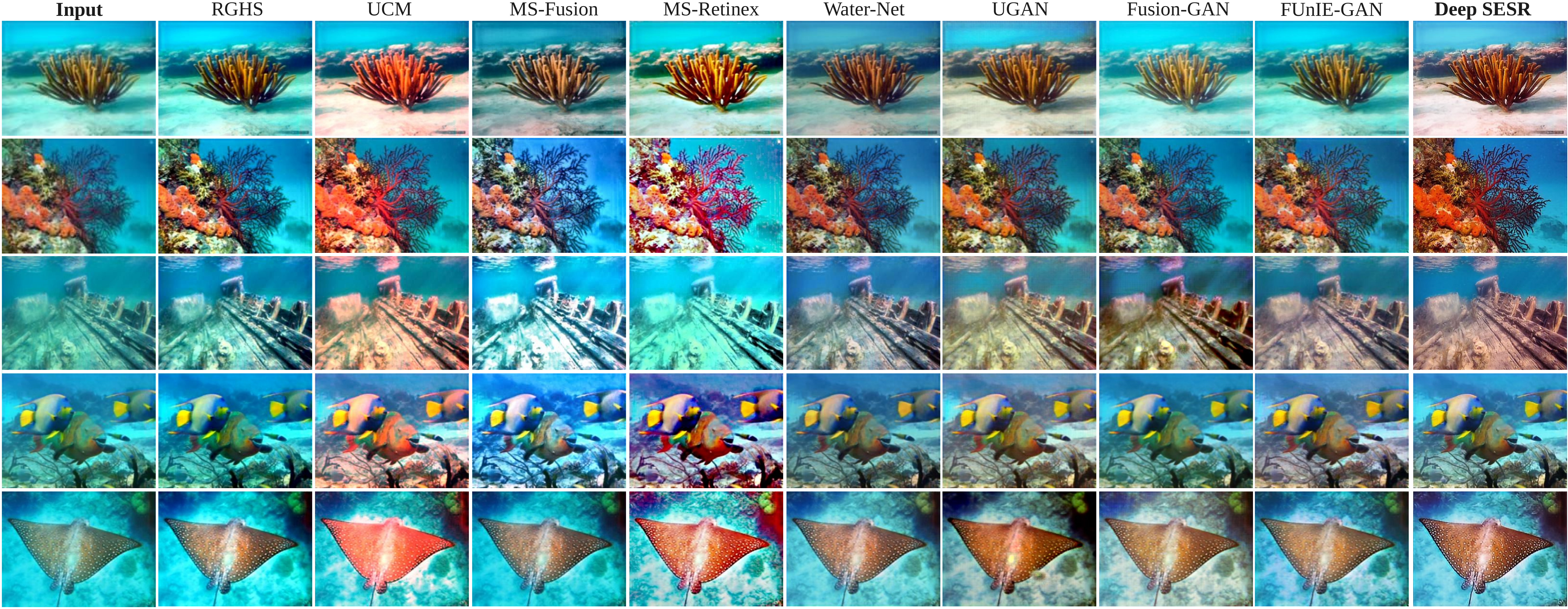}%
        \vspace{-1mm}
        \caption{Qualitative comparison of Deep SESR-enhanced images with SOTA models: RGHS~\cite{huang2018shallow}, UCM~\cite{iqbal2010enhancing}, MS-Fusion~\cite{ancuti2012enhancing}, MS-Retinex~\cite{zhang2017underwater}, Water-Net~\cite{li2019underwater}, UGAN~\cite{fabbri2018enhancing}, Fusion-GAN~\cite{li2019fusion}, and FUnIE-GAN~\cite{islam2019fast}.}%
        \label{fig:quan_en_comp}
        \vspace{1mm}
\end{figure*}

\begin{table*}[t]
\centering
\caption{Quantitative performance comparison for super-resolution: scores are shown as $\textit{mean} \pm \sqrt{\textit{variance}}$; the first and second best scores (in each column per-dataset) are colored red, and blue, respectively. ($^\ddagger$Does not support $3\times$ scale)}
%\footnotesize
\scriptsize
\vspace{-1mm}
\begin{tabular}{c|l||c|c|c||c|c|c||c|c|c}
  \hline
   & & \multicolumn{3}{c||}{\textit{PSNR}} & \multicolumn{3}{c||}{\textit{SSIM}} & \multicolumn{3}{c}{\textit{UIQM}} \\  
  \cline{2-11} 
  & Model & $2\times$ & $3\times$ & $4\times$ & $2\times$ & $3\times$ & $4\times$ & $2\times$ & $3\times$ & $4\times$ \\ \hline
  \hline
  \parbox[t]{1mm}{\multirow{6}{*}{\rot{\textbf{UFO-120}}}} & SRCNN & $24.75 \pm 3.7$ & $22.22 \pm 3.9$   & $19.05 \pm 2.3$  & $.72 \pm .07$  & $.65 \pm .09$  & $.56 \pm .12$  & $2.39 \pm 0.35$  & $2.24 \pm 0.17$  & $2.02 \pm 0.47$    \\ 
  & SRResNet & $25.23 \pm 4.1$  & $23.85 \pm 2.8$  & $19.13 \pm 2.4$  & $.74 \pm .08$  &  $.68 \pm .07$ & $.56 \pm .05$  & $2.42 \pm 0.37$  & $2.18 \pm 0.26$  &  $2.09 \pm 0.30$   \\ 
  & SRGAN & $26.11 \pm 3.9$  & {\color{blue}$23.87 \pm 4.2$}  & $21.08 \pm 2.3$  & $.75 \pm .06$  & $.70 \pm .05$  & $.58 \pm .09$  & $2.44 \pm 0.28$  & {\color{blue}$2.39 \pm 0.25$}  &  $2.26 \pm 0.17$   \\ 
  & RSRGAN & $25.25 \pm 4.3$  & $23.15 \pm 4.1$  & $20.25 \pm 2.4$  & {\color{blue}$.79 \pm .08$}  & {\color{blue}$.71 \pm .08$} & $.58 \pm .04$  & $2.41 \pm 0.29$  & $2.38 \pm 0.31$  &  $2.27 \pm 0.22$   \\ 
  & SRDRM$^\ddagger$ & $26.23 \pm 4.4$ & $-$  & {\color{blue}$22.26 \pm 2.5$}  & {\color{blue}$.79 \pm .09$}  &  $-$ & {\color{blue}$.59 \pm .05$}  & {\color{blue}$2.45 \pm 0.43$}  & $-$  & {\color{blue}$2.28 \pm 0.35$}  \\ 
  & SRDRM-GAN$^\ddagger$ & {\color{blue}$26.26 \pm 4.3$} & $-$ & $22.21 \pm 2.4$  & $.78 \pm .08$  &  $-$ & $.58 \pm .13$  & $2.42 \pm 0.30$  & $-$  &  $2.27 \pm 0.44$   \\ 
  & Deep SESR & {\color{red}$28.57 \pm 3.5$}  & {\color{red}$26.86 \pm 4.1$}  & {\color{red}$24.75 \pm 2.8$}  & {\color{red}$.85 \pm .09$}  & {\color{red}$.75 \pm .06$}  & {\color{red}$.66 \pm .05$}  & {\color{red}$3.09 \pm 0.41$}  & {\color{red}$2.87 \pm 0.39$}  & {\color{red}$2.55 \pm 0.35$}  \\ \hline
%\parbox[t]{1mm}{\multirow{6}{*}{\rot{\textbf{UFO: Set-F}}}} & SRDRM &   & $-$  &   &   &  $-$ &   &   & $-$  &     \\ 
%  & SRDRM-GAN &   &   &   &   &   &   &   &   &     \\ 
%  & RSRGAN &   &   &   &   &   &   &   &   &     \\ 
%  & SRGAN &   &   &   &   &   &   &   &   &     \\ 
%  & SRCNN &   &   &   &   &   &   &   &   &     \\
%  & Deep SESR & $18.85 \pm 1.7$  &  $18.85 \pm 1.7$ &  $18.85 \pm 1.7$ &  $.77 \pm .06$ &  $.77 \pm .06$ &  $.77 \pm .06$ &  $2.20 \pm 0.69$ &  $2.20 \pm 0.69$ &  $2.20 \pm 0.69$   \\ \hline
%\parbox[t]{1mm}{\multirow{6}{*}{\rot{\textbf{UFO: Set-O}}}} & SRDRM &   & $-$  &   &   & $-$  &   &   &  $-$ &     \\ 
%  & SRDRM-GAN &   &   &   &   &   &   &   &   &     \\ 
%  & RSRGAN &   &   &   &   &   &   &   &   &     \\ 
%  & SRGAN &   &   &   &   &   &   &   &   &     \\ 
%  & SRCNN &   &   &   &   &   &   &   &   &     \\ 
%  & Deep SESR & $18.85 \pm 1.7$  &  $18.85 \pm 1.7$ &  $18.85 \pm 1.7$ &  $.77 \pm .06$ &  $.77 \pm .06$ &  $.77 \pm .06$ &  $2.20 \pm 0.69$ &  $2.20 \pm 0.69$ &  $2.20 \pm 0.69$   \\ \hline
\parbox[t]{1mm}{\multirow{6}{*}{\rot{\textbf{USR-248}}}} & SRCNN & $24.88 \pm 4.4$  & $24.01 \pm 3.5$   & $23.75 \pm 3.2$  &  $.73 \pm .08$ & $.70 \pm .10$  & $.69 \pm .12$  & $2.38 \pm 0.38$  & $2.31 \pm 0.29$  &  $2.21 \pm 0.68$   \\ 
  & SRResNet &  $24.96 \pm 3.7$ & $23.39 \pm 5.2$  & $22.21 \pm 3.6$  &  $.74 \pm .07$ & $.71 \pm .11$  & $.70 \pm .08$  &  $2.42 \pm 0.48$ & $2.33 \pm 0.58$  & $2.27 \pm 0.70$    \\ 
  & SRGAN & $25.76 \pm 3.5$  & {\color{blue}$25.02 \pm 3.9$}  & $24.36 \pm 4.3$  &  $.77 \pm .06$ & {\color{blue}$.75 \pm .05$}  & $.69 \pm .13$  & $2.53 \pm 0.42$  &  {\color{blue}$2.65 \pm 0.44$} & $2.75 \pm 0.66$\\ 
  & RSRGAN & $25.11 \pm 2.9$  &  $24.96 \pm 4.7$  &   $24.15 \pm 2.9$ & $.75 \pm .06$  & $.72 \pm .09$  &  {\color{blue}$.71 \pm .09$} & $2.42 \pm 0.35$  & $2.49 \pm 0.56$  &  $2.55 \pm 0.47$    \\ 
  & SRDRM$^\ddagger$ & $26.16 \pm 3.5$  &  $-$ & {\color{red}$24.96 \pm 3.3$}  & $.77 \pm .10$  & $-$  &  {\color{red}$.72 \pm .11$} & $2.47 \pm 0.69$  &  $-$ & $2.35 \pm 0.51$     \\ 
  & SRDRM-GAN$^\ddagger$ & {\color{blue}$26.77 \pm 4.1$}  & $-$  & {\color{blue}$24.77 \pm 3.4$}  & {\color{blue}$.82 \pm .07$}  & $-$  & $.70 \pm .12$  & {\color{blue}$2.87 \pm 0.55$}  &  $-$ & {\color{blue}$2.81 \pm 0.56$}     \\ 
  & Deep SESR & {\color{red}$27.03 \pm 2.9$}  & {\color{red}$25.92 \pm 3.5$}  &  $24.59 \pm 3.8$ & {\color{red}$.88 \pm .05$}  & {\color{red}$.76 \pm .05$}  & {\color{blue}$.71 \pm .08$}  & {\color{red}$3.15 \pm 0.44$}  & {\color{red}$3.04 \pm 0.37$}  & {\color{red}$2.96 \pm 0.28$}  \\ \hline
\end{tabular}
\label{tab:quan_sr}
\vspace{2mm}
\end{table*}

\begin{figure*}[t]
    \centering
    \includegraphics[width=\linewidth]{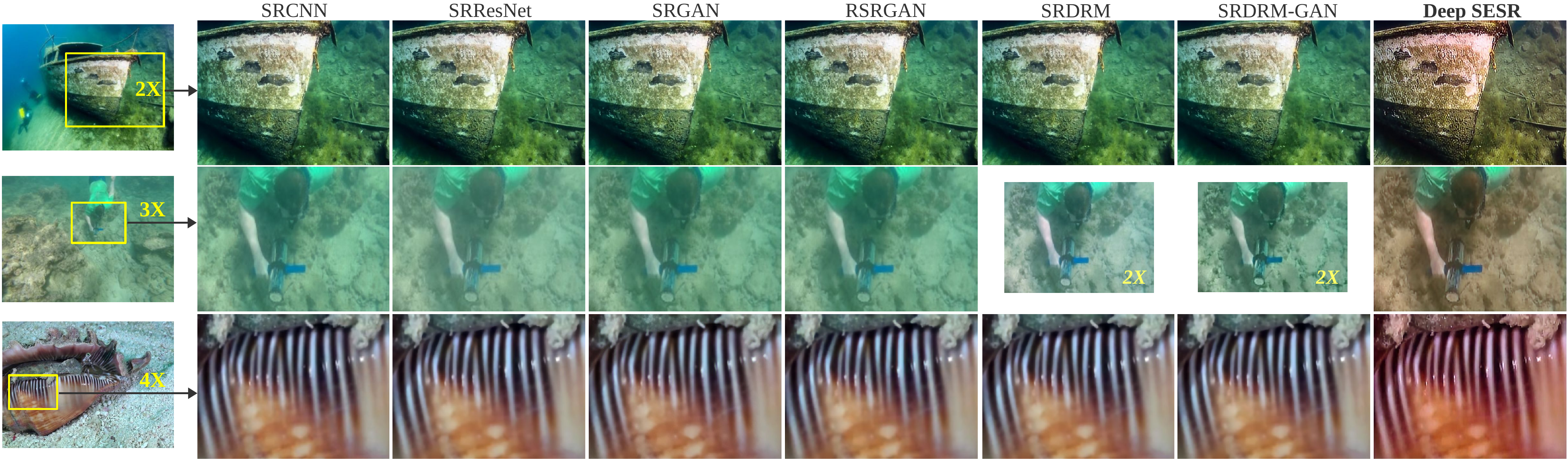}%
      \vspace{-1mm}
        \caption{Qualitative comparison for SISR performance of Deep SESR with existing solutions and SOTA models: SRCNN~\cite{dong2015image}, SRResNet~\cite{ledig2017photo}, SRGAN~\cite{ledig2017photo}, RSRGAN~\cite{chen2019recovering}, SRDRM~\cite{islam2019underwater}, and SRDRM-GAN~\cite{islam2019underwater}.}%
    \label{fig:comp_sesr}
    \vspace{2mm}
\end{figure*}

\section{Experimental Results}
\subsection{Implementation Details}\label{implement}
As mentioned in Section~\ref{sec:data_prep}, Deep SESR training is supervised by paired data of the form $(\{X\}, \{S, E, Y\})$. We use TensorFlow libraries~\cite{abadi2016tensorflow} to implement the optimization pipeline (of Eq.~\ref{eq:final}); a Linux host with two Nvidia\texttrademark{ }GTX 1080 graphics cards are used for training. Adam optimizer~\cite{kingma2014adam} is used for the global iterative learning with a rate of $10^{-4}$ and a momentum of $0.5$; the network converges within $23$-$26$ epochs of training in this setup (with a batch-size of $2$). In the following sections, we present the experimental results based on qualitative analysis, quantitative evaluations, and ablation studies. Since there are no existing SESR methods, we compare the Deep SESR performance separately with SOTA image enhancement and super-resolution models. Note that, all models in comparison are trained on the same train-validation splits (of respective datasets) by following their recommended parameter settings. Also, for datasets other than UFO-120, the AAN (and $\mathcal{L}_{Contrast}^{LR}$) is not used by Deep SESR as their ground truth saliency maps are not available.

%\begin{table}[h]
%\centering
%\caption{Dataset-specific information and hyper-parameters used by Deep SESR model for training and validation.}
%\footnotesize
%\vspace{-1mm}
%\begin{tabular}{l||c|c|c}
%  \hline
%  \textbf{Dataset} & Train/validation (\#) & Epochs & Batch-size \\ \hline \hline
%  UFO &  $1500/150$ &  $30$   & $2$  \\
%  EUVP (Paired)~\cite{islam2019fast} &  $6000/600$ &  $25$   & $8$  \\
%  UImNet~\cite{fabbri2018enhancing} &  $4260/400$ &  $10$   & $8$  \\
%  USR-248~\cite{islam2019underwater} &  $2400/240$ &  $20$   & $4$  \\
%  \hline
 % U45~\cite{li2019fusion} & \multicolumn{3}{c}{Only for testing (\#: 45)}  \\ %\hline
  %Set5/Set14/Sun80~\cite{bevilacqua2012low,zeyde2010single,sun2012super} & \multicolumn{3}{c}{*Terrestrial data for testing (\#: 5/14/80)}  \\ \hline
%\end{tabular}
%\vspace{-3mm}
%\label{tab:train}
%\end{table}%

\subsection{Evaluation: Enhancement}
We first qualitatively analyze the Deep SESR-generated images in terms of color, contrast, and sharpness. As Fig.~\ref{fig:qual_en} illustrates, the enhanced images are perceptually similar to the respective ground truth. Specifically, the greenish underwater hue is rectified, true pixel colors are mostly restored, and the global image sharpness is recovered. 
Moreover, the generated saliency map suggests that it focused on the right foreground regions for contrast improvement. We further demonstrate the contributions of each loss-term: $\mathcal{L}_{Contrast}^{LR}$, $\mathcal{L}_{P}^{LR}$, $\mathcal{L}_{Color}^{LR}$, and $\mathcal{L}_{Content}^{LR}$ for learning the enhancement. 
We observe that the color rendition gets impaired without $\mathcal{L}_{P}^{LR}$ and $\mathcal{L}_{Color}^{LR}$, whereas, $\mathcal{L}_{Content}^{LR}$ contributes to learning finer texture details. We also notice a considerably low-contrast image generation without $\mathcal{L}_{Contrast}^{LR}$, which validates the utility of saliency-driven contrast evaluation via CMI (see Section~\ref{loss_fun}).

Next, we compare the perceptual image enhancement performance of Deep SESR with the following models: (i) relative global histogram stretching (RGHS)~\cite{huang2018shallow}, (ii) unsupervised color correction (UCM)~\cite{iqbal2010enhancing}, (iii) multi-scale fusion (MS-Fusion)~\cite{ancuti2012enhancing}, (iv) multi-scale Retinex (MS-Retinex)~\cite{zhang2017underwater}, (v) Water-Net~\cite{li2019underwater}, (vi) UGAN~\cite{fabbri2018enhancing}, (vii) Fusion-GAN~\cite{li2019fusion}, and (viii) FUnIE-GAN~\cite{islam2019fast}. The first four are physics-based models and the rest are learning-based models; they provide SOTA performance for underwater image enhancement in RGB space (without requiring scene depth or optical waterbody measures). Their performance is quantitatively evaluated on common test sets of each dataset based on standard metrics~\cite{islam2019fast,liu2019real}: peak signal-to-noise ratio (PSNR)~\cite{hore2010image}, structural similarity measure (SSIM)~\cite{wang2004image}, and underwater image quality measure (UIQM)~\cite{panetta2016human}. The PSNR and SSIM quantify reconstruction quality and structural similarity of generated images (with respect to ground truth), whereas the UIQM evaluates image qualities based on colorfulness, sharpness, and contrast. 
The evaluation is summarized in Table~\ref{tab:quan_se}; moreover, a few qualitative comparisons are shown in Fig.~\ref{fig:quan_en_comp}.

As Fig.~\ref{fig:quan_en_comp} demonstrates, UCM and MS-Retinex often suffer from over-saturation, whereas RGBH, MS-Fusion, and Water-Net fall short in hue rectification. In comparison, the color restoration and contrast enhancement of UGAN, Fusion-GAN, and FUnIE-GAN are generally better. In addition to achieving comparable color recovery and hue rectification, the Deep SESR-generated images are considerably sharper. Since the boost in performance is rather significant for UFO-120 dataset (suggested by the results of Table~\ref{tab:quan_se}), it is likely that the additional knowledge about foreground pixels through $\mathcal{L}_{Contrast}^{LR}$ helps in this regard. Deep SESR achieves competitive and often better performance in terms of PSNR and SSIM as well. In particular, it generally attains better UIQM scores; we postulate that $\mathcal{L}_{P}^{LR}$ contributes to this enhancement, as it is designed to improve the UICM (see Section~\ref{loss_fun}). Further ablation investigations reveal a $9.47\%$ drop in UIQM values without using $\mathcal{L}_{P}^{LR}$ in the learning objective.

\begin{table}[h]
%\vspace{-1mm}
%\footnotesize
\scriptsize
    \centering
    \caption{Deep SESR performance on the UFO-120 dataset; set-wise mean scores are shown for $2\times$/$3\times$/$4\times$ SESR.}
    \begin{tabular}{l||c|c|c}
    \hline
      & \textit{PSNR} & \textit{SSIM} & \textit{UIQM} \\ \hline %\hline  
        \textbf{Set-U} & $28.55/26.77/24.25$ & $0.86/0.75/0.66$ & $3.07/2.89/2.54$ \\
        \textbf{Set-F} & $27.93/26.33/24.87$ & $0.85/0.73/0.63$ & $3.10/2.84/2.52$ \\
        \textbf{Set-O} & $28.95/27.15/25.45$ & $0.84/0.79/0.68$ & $3.09/2.86/2.58$ \\ \hline
    \end{tabular}
    \label{tab:ufo_quant}
    \vspace{1mm}
\end{table}

\begin{figure}[h]
%\vspace{1mm}
    \centering
    \includegraphics[width=\linewidth]{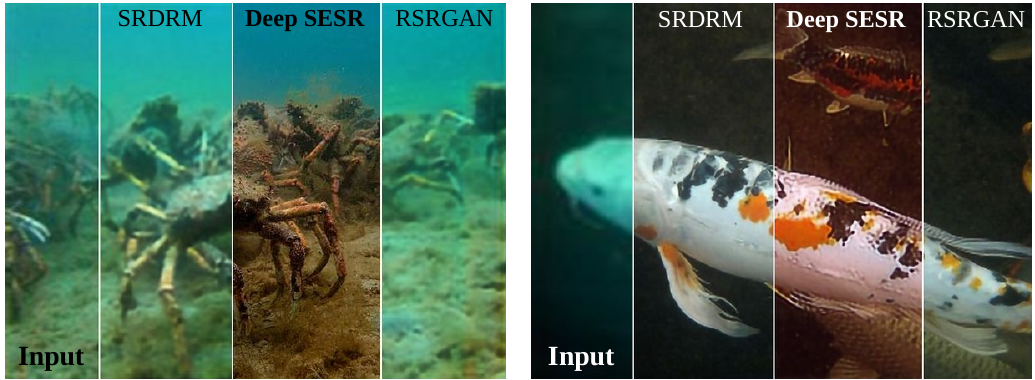}%
    \vspace{-1mm}
        \caption{Color and texture recovery of Deep SESR: comparison shown with two best-performing SISR models (as of Table~\ref{tab:quan_sr}).}%
    \label{fig:qual_sr}
    \vspace{2mm}
\end{figure}

\subsection{Evaluation: Super-Resolution}
We follow similar experimental procedures for evaluating the super-resolution performance of Deep SESR. We consider the existing underwater SISR models named RSRGAN~\cite{chen2019recovering}, SRDRM~\cite{islam2019underwater}, and SRDRM-GAN~\cite{islam2019underwater} for performance comparison. 
We also include the standard (terrestrial) SISR models named SRCNN~\cite{dong2015image}, SRResNet~\cite{ledig2017photo}, and SRGAN~\cite{ledig2017photo} in the evaluation as benchmarks. We compare their $2\times$, $3\times$, and  $4\times$ SISR performance on two large-scale datasets: UFO-120, and USR-248. The results are presented in Table~\ref{tab:quan_sr}, and a few samples are shown in Fig.~\ref{fig:comp_sesr}. Note that, the test images of USR-248 dataset are left undistorted for a fair comparison.   
%we only train the main branch of Deep SESR (\ie, with $\lambda_{c}^{LR}=\lambda_{f}^{LR}=\lambda_{t}^{LR}=0$) for a fair comparison on the USR-248 dataset.          

As Table~\ref{tab:quan_sr} demonstrates, Deep SESR outperforms other models in comparison by considerable margins on UIQM. This is due to the fact that it enhances perceptual image qualities in addition to spatial resolution. As shown in Fig.~\ref{fig:comp_sesr}, Deep SESR generates much sharper and better quality HR images from both distorted and undistorted LR input patches, which contributes to its competitive PSNR and SSIM scores on the USR-248 dataset. 
Fig.~\ref{fig:qual_sr} further demonstrates that it does not introduce noise by unnecessary over-correction, which is a prevalent limitation of existing automatic image enhancement solutions.
Lastly, we observe similar performance trends for all three types of spatial down-sampling, \ie, for Set-U, Set-F, and Set-O (see Section~\ref{sec:data_prep}); we present the relative quantitative scores in Table~\ref{tab:ufo_quant}.

\begin{figure}[t]
%%\vspace{-1mm}
    \centering
    \begin{subfigure}{0.5\textwidth}
    \centering
        \includegraphics[width=\linewidth]{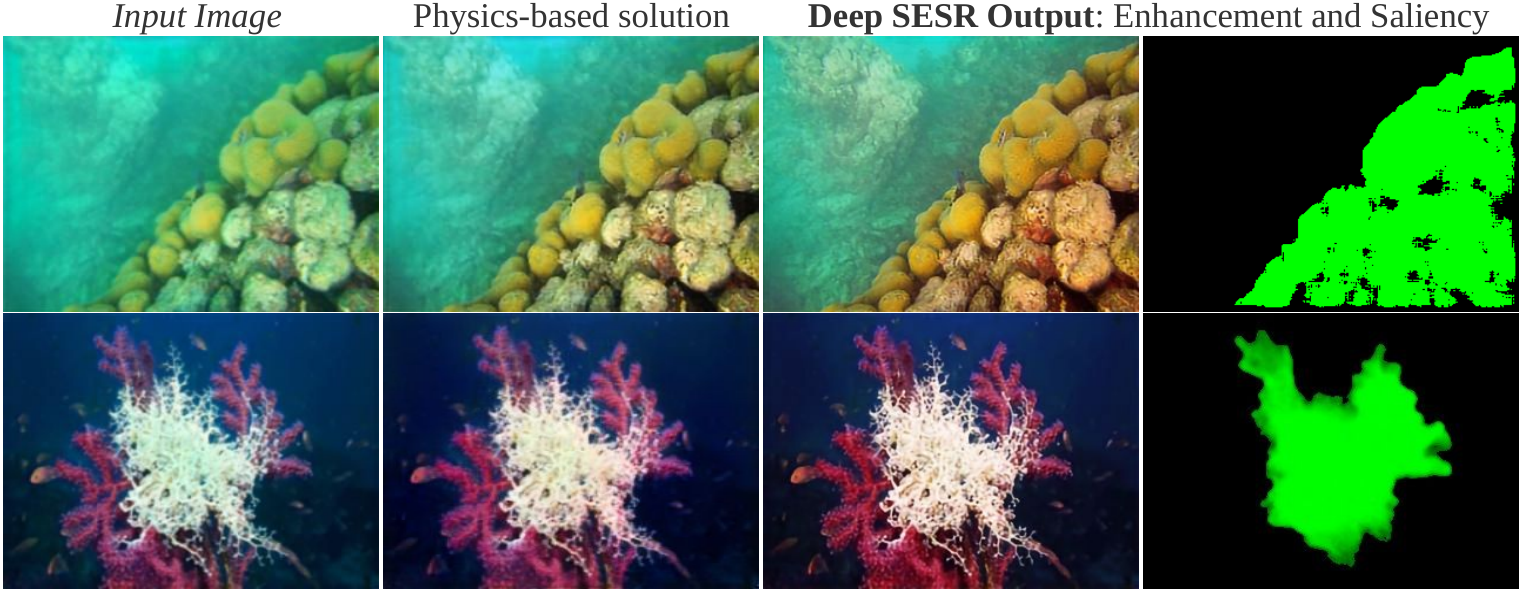}%
        %\vspace{-1mm}
        \caption{Comparison with a physics-based color restoration method~\cite{berman2018underwater} that uses spectral waterbody measures and haze-lines prior.}%
        \label{fig:qual_ph}
    \end{subfigure}
    \vspace{3mm}
    
    \begin{subfigure}{0.5\textwidth}
        \centering
       \includegraphics[width=\linewidth]{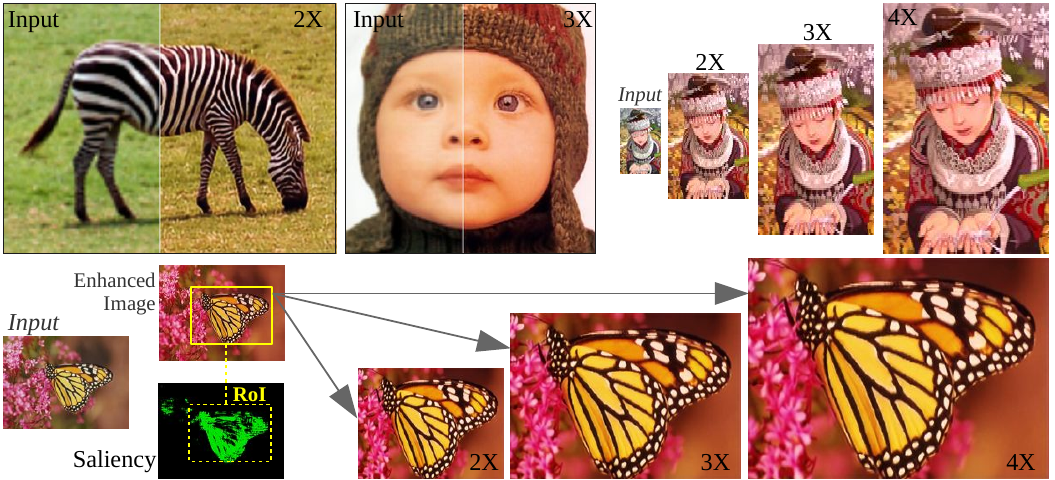}%
        %\vspace{-1mm}
        \caption{Performance for $2\times$, $3\times$, and $4\times$ SESR on terrestrial images.}%
        \label{fig:qual_terr}
    \end{subfigure}
    %\vspace{-1mm}
    \caption{Demonstration of generalization performance of Deep SESR model (trained on UFO-120 dataset).}%
    \label{fig:qual_gen}
    %\vspace{-4mm}
\end{figure}

\section{Generalization Performance}
Due to the ill-posed nature of modeling underwater image distortions without scene-depth and optical waterbody measurements, learning-based solutions often fail to generalize beyond supervised data. In addition to the already-presented results, we demonstrate the color and texture recovery of Deep SESR on unseen natural images in Fig.~\ref{fig:qual_gen}. As seen in Fig.~\ref{fig:qual_ph}, Deep SESR-enhanced pixel intensities are perceptually similar to a comprehensive physics-based approximation~\cite{berman2018underwater}. Additionally, it generates the respective HR images and saliency maps, and still offers more than $10$ times faster run-time.

Deep SESR also provides reasonable performance on terrestrial images. As demonstrated in Fig.~\ref{fig:qual_terr}, the color and texture enhancement of unseen objects (\eg, grass, face, clothing, etc.) are perceptually coherent. Moreover, as Table~\ref{tab:terr_quant} indicates, its performance in terms of sharpness and contrast recovery for $2\times$, $3\times$, and $4\times$ SISR are competitive with SOTA benchmark results~\cite{li2019feedback,zhang2018residual}. Note that, much-improved performance can be achieved by further tuning and training on terrestrial datasets. Nevertheless, these results validate that the proposed architecture has the capacity to learn a generalizable solution of the underlying SESR problem. 
%The PSNR scores are slightly lower, which is likely due to the additional color enhancement performed by Deep SESR (as PSNR is calculated based on mean-squared error in pixel intensities).             

\begin{table}[h]
\footnotesize
%\scriptsize
    \centering
    \caption{Deep SESR performance on terrestrial test data; blue (and boldfaced) scores represent $3\%$ (and $1\%$) margins with SOTA benchmark results for $2\times$/$3\times$/$4\times$ SISR~\cite{li2019feedback,zhang2018residual}.}%
    \vspace{-1mm}
    \begin{tabular}{l||c|c}
    \hline
      & \textit{PSNR} & \textit{SSIM} \\ \hline %\hline  
        \textbf{Set5}~\cite{bevilacqua2012low} & $29.87$ / {\color{blue}$28.77$} / ${\color{blue}26.14}$ & ${\color{blue}0.925}$ / ${\color{blue}\mathbf{0.908}}$ / ${\color{blue}0.855}$  \\
        \textbf{Set14}~\cite{zeyde2010single} & {\color{blue}$28.78$} / $27.34$ / ${\color{blue}26.89}$ & ${\color{blue}\mathbf{0.914}}$ / ${\color{blue}0.801}$ / ${\color{blue}0.756}$  \\
        \textbf{Sun80}~\cite{sun2012super} & $25.73$ / $23.18$ / $21.05$ & ${\color{blue}\mathbf{0.802}}$ / ${\color{blue}0.755}$ / $0.704$  \\ \hline
    \end{tabular}
    \label{tab:terr_quant}
    \vspace{-3mm}
\end{table}

\section{Operational Feasibility \& Design Choices}
\begin{figure}[h]
%\vspace{-1mm}
    \centering
        \includegraphics[width=0.99\linewidth]{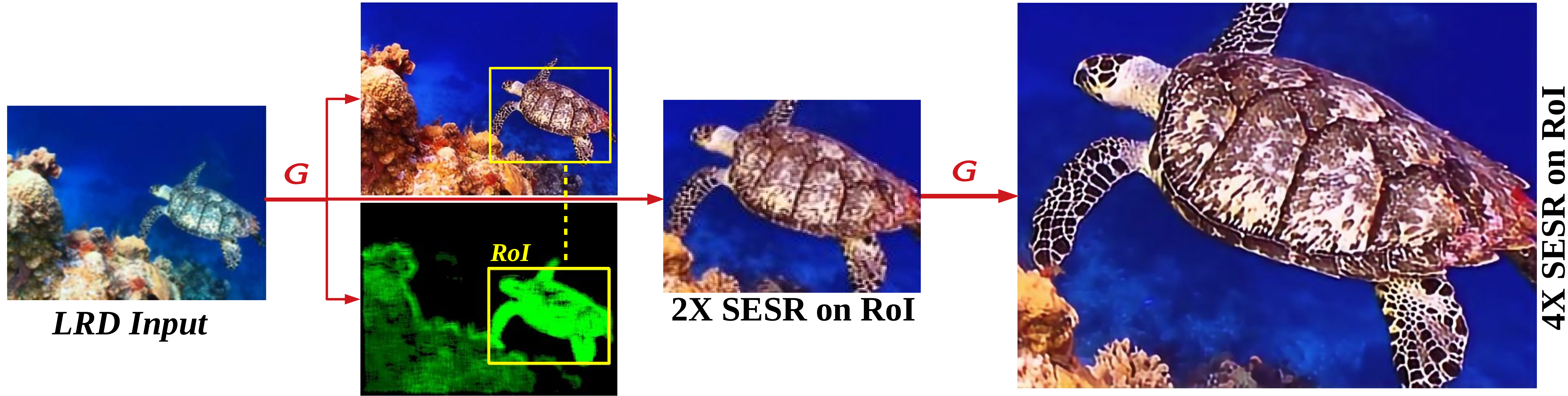} 
        \vspace{-1mm}
        \caption{Demonstration of automatic RoI selection based on local intensity values in the saliency map; Deep SESR can be applied again on the enhanced RoI for a detailed perception.}
        \vspace{-1mm}
    \label{fig:qual_sal}
    %\vspace{-1mm}
\end{figure}

Deep SESR's on-board memory requirement is only $10$ MB, and it offers a run-time of $129$ milliseconds (ms) per-frame, \ie, $7.75$ frames-per-second (FPS) on a single-board computer: Nvidia\texttrademark AGX Xavier. 
As shown in Table~\ref{tab:time}, it provides much faster speeds for the following design choices: 

\textit{1)} \textbf{Learning $\hat{E}$ and $\hat{S}$ on separate branches} facilitates a faster run-time when HR perception is not required. Specifically, we can decouple the $X$ $\rightarrow$ $S$, $E$ branches from the frozen model, which operates at $10.02$ FPS ($22\%$ faster) to perform enhancement and saliency prediction. As shown in Fig.~\ref{fig:qual_sal}, the predicted saliency map can be exploited for automatic RoI selection by using density gradient estimation techniques such as mean-shift~\cite{comaniciu1999mean}. The SESR output corresponding to the RoI can be generated with an additional $25$ ms of processing time.

\begin{table}[h]
%\vspace{-1mm}
\centering
\caption{Run-time comparison for various design choices of Deep SESR (on Nvidia\texttrademark AGX Xavier).}
\vspace{-1mm}
\footnotesize
\begin{tabular}{l|c|c}
  \hline
   & $X$ $\rightarrow$ $S$, $E$  & $X$ $\rightarrow$ $S$, $E$ , $Y$   \\ \hline 
  \textbf{With FENet-1d} &  $87.3$ ms ($11.45$ FPS)  &  $113$ ms ($8.85$ FPS)   \\
    \textbf{With FENet-2d} &  $99.8$ ms ($10.02$ FPS)  &  $129$ ms ($7.75$ FPS)   \\ \hline
\end{tabular}
%\vspace{-4mm}
\label{tab:time}
\end{table}%

\textit{2)} \textbf{FENet-1d and FENet-2d} are two design choices for the FENet (see Fig.~\ref{fig:model_fnet}); FENet-2d is the default architecture that learns $3\times3$ and $5\times5$ filters in two parallel branches, whereas, FENet-1d refers to using a single branch of $3\times3$ filters. As shown in Table~\ref{tab:time}, faster feature extraction by FENet-1d facilitates a $12.5\%$ speed-up for Deep SESR. 
However, we observe a slight drop in performance, \eg, $1.8\%$/$1.5\%$/$1.8\%$ lower scores for PSNR/SSIM/UIQM on UFO-120 dataset. Nevertheless, the generated images are qualitatively indistinguishable and the trade-off is admissible in practical applications.

Overall, Deep SESR offers use-case-specific design choices and ensures computational efficiency with robust SESR performance. These features make it suitable for near real-time robotic deployments; further demonstration is available in the supplementary material.  
\section{Conclusion}
 In this paper, we introduce the problem of simultaneous enhancement and super-resolution (SESR) and present an efficient learning-based solution for underwater imagery. The proposed generative model, named Deep SESR, can learn $2\times$$-$$4\times$ SESR and saliency prediction on a shared feature space. We also present its detailed network architecture, associated loss functions, and end-to-end training pipeline. Additionally, we contribute over $1500$ annotated samples to facilitate large-scale SESR training on the UFO-120 dataset. We perform a series of qualitative and quantitative experiments, which suggest that Deep SESR: \textit{i)} provides SOTA performance on underwater image enhancement and super-resolution, \textit{ii)} exhibits significantly better generalization performance on natural images than existing solutions, \textit{iii)} provides competitive results on terrestrial images, and \textit{iv)} achieves fast inference on single-board platforms. 
 The inspiring performance, computational efficiency, and availability of application-specific design choices make Deep SESR suitable for near real-time use by visually-guided underwater robots. 
In the future, we seek to incorporate $6\times$$-$$8\times$ spatial upscaling capability into the model with reasonable performance trade-offs.

\section*{Acknowledgements}
We acknowledge the support of the MnDrive initiative at the University of Minnesota\footnote{\url{https://mndrive.umn.edu/}} for our research. 
We are also grateful to the Bellairs Research Institute\footnote{\url{https://www.mcgill.ca/bellairs/}} of Barbados for providing us with the facilities for field experiments. 
Additionally, we thank Nvidia\texttrademark{ }for donating two GPUs for our work.  
Finally, we acknowledge our colleagues at the IRVLab\footnote{\url{http://irvlab.cs.umn.edu/}} for their assistance in data collection, annotation, and in preparation of media files.

{\small
\bibliographystyle{ieee}
\bibliography{egbib}
}

\section*{Appendix I: Dataset Information}
{\small

\begin{itemize}
    \item The UFO-120 dataset: \url{http://irvlab.cs.umn.edu/resources/ufo-120-dataset}.
    
    \item The USR-248~\cite{islam2019underwater} dataset: \url{http://irvlab.cs.umn.edu/resources/usr-248-dataset}.
        
    \item The EUVP~\cite{islam2019fast} and UImNet~\cite{fabbri2018enhancing} datasets: \\ \url{http://irvlab.cs.umn.edu/resources/euvp-dataset}.
    
    %\item The U45~\cite{li2019fusion} dataset: \\ \url{https://github.com/IPNUISTlegal/underwater-test-dataset-U45-}
    
    \item The Set5~\cite{bevilacqua2012low},  Set14~\cite{zeyde2010single}, Sun80~\cite{sun2012super}, and other terrestrial datasets: \url{https://github.com/ChaofWang/Awesome-Super-Resolution}.
    
\end{itemize}

}

\section*{Appendix II: Credits for Media Resources}
{\small
\begin{enumerate}   
    \item David Gantt. A Sea Fan at Karpata. 2012. (Flickr): \\ \url{https://www.flickr.com/photos/i2image/23681523755/}.

    \item SeaPics.com. Diver Over a Reef.  2015. (SeaPics): \\
    \url{https://images.seapics.com/images/LowRez/0043001/}.

    \item Vincent POMMEYROL. Underwater Photography. 2012. (Flickr): 
    \url{https://www.flickr.com/photos/vincentpommeyrol/8301150254/}.

    \item BelleDeesse. Watercraft. 2013. (WallpaperUP): \\ \url{https://www.wallpaperup.com/72705/Boat_Underwater_Ocean_Diver.html}.
    
    \item Nick Shaw. Leptomithrax gaimardii (Great Spider Crab). 2018. (Atlas of Living Australia): 
    \url{https://images.ala.org.au/image/details?imageId=8b7d6bad-1ffb-4740-8161-d3c6d4f35018}.
    
    \item David Piano. City of Grand Rapids Shipwreck watermarked-2. 2018. (Flickr):
    \url{https://www.flickr.com/photos/davidpiano/44403874204/}
    
    \item Koi Photos Underwater View. The Pond Experts. 2014. (PondExperts):
    \url{https://www.pondexperts.ca/wp-content/uploads/2014/07/img_3870.jpg}
    
    \item Cat Trumpet. 2 Hours of Beautiful Coral Reef Fish, Relaxing Ocean Fish, 1080p HD. 2016. (YouTube): \\ \url{https://youtu.be/cC9r0jHF-Fw}.

    \item Nature Relaxation Films. 3 Hours of Stunning Underwater Footage, French Polynesia, Indonesia. 2018. (YouTube): \\
    \url{https://youtu.be/eSRj847AY8U}.

    \item Calm Cove Club - Relaxing Videos. 4K Beautiful Ocean Clown Fish Turtle Aquarium. 2017 (YouTube): \\ \url{https://youtu.be/DP4QDNm6f4Q}.
    
    \item Scubasnap.com. 4K Underwater at Stuart Cove's, 2014 (YouTube):  
    \url{https://youtu.be/kiWfG31YbXo}.

    \item 4.000 PIXELS. Beautiful Underwater Nature. 2017 (YouTube): \url{https://youtu.be/1-Cn0b1MKrM}.
    
    \item Magnus Ryan Diving. SCUBA Diving Egypt Red Sea. 2017 (YouTube):
    \url{https://youtu.be/CaLfMHl3M2o}.
    
    %\item Soothing Relaxation. Sleep Music in Underwater Paradise. 2017 (YouTube): \url{https://youtu.be/OVct34NUk3U}.

    % all the other videos
    %% Fishes and reefs
    %\item Amby-Relaxing Music. Stunning Aquarium. 2017. (YouTube): \url{https://youtu.be/-j2uKUK5P10}.
    
    %\item Yellow Brick Cinema - Relaxing Music. Relaxing Piano Music Video. 2019. (YouTube): \url{https://youtu.be/gqklulvlTls}.
    
    \item TheSilentWatcher. 4K Coral World-Tropical Reef. 2018. (YouTube):
    \url{https://youtu.be/uyb0wW0ln_g}.
    
    \item Awesome Video. 4K- The Most Beautiful Coral Reefs and Undersea Creature on Earth. 2017. (YouTube): \\
    \url{https://youtu.be/nvq_lvC1MRY}.
    
    \item Earth Touch. Celebrating World Oceans Day in 4K. 2015. (YouTube):
    \url{https://youtu.be/IXxfIMNgMJA}.
    
    \item BBC Earth. Deep Ocean: Relaxing Oceanscapes. 2018. (YouTube):
    \url{https://youtu.be/t_S_cN2re4g}.
    
    \item Alegra Chetti. Let's Go Under the Sea I Underwater Shark Footage I Relaxing Underwater Scene. 2016. (YouTube):
    \url{https://youtu.be/rQB-f5BHn5M}.
    
    \item Underwater 3D Channel- Barry Chall Films. Planet Earth, The Undersea World (4K). 2018. (YouTube): \\
    \url{https://youtu.be/567vaK3BKbo}.
    
    \item Undersea Productions. ``ReefScapes: Nature's Aquarium" Ambient Underwater Relaxing Natural Coral Reefs and Ocean Nature. 2009. (YouTube): \\
    \url{https://youtu.be/muYaOHfP038}.
    
    \item BBC Earth. The Coral Reef: 10 Hours of Relaxing Oceanscapes. 2018. (YouTube): \\
    \url{https://youtu.be/nMAzchVWTis}.
    
    \item Robby Michaelle. Scuba Diving the Great Barrier Reef Red Sea Egypt Tiran. 2014. (YouTube): \\
    \url{https://youtu.be/b7BEAsyPgHM}.
    
    \item Bubble Vision. Diving in Bali. 2012. (YouTube): \\
    \url{https://youtu.be/uCRBxtQ55_Y}. 
    
    \item Vic Stefanu - Amazing World Videos. EXPLORING The GREAT BARRIER REEF, fantastic UNDERWATER VIDEOS (Australia). 2015. (YouTube): \\
    \url{https://youtu.be/stMzgmPlQQM}.
    
    \item Our Coral Reef. Breathtaking Dive in Raja Ampat, West Papua, Indonesia Coral Reef. 2018. (YouTube): \\
    \url{https://youtu.be/i4ZSMDWNXTg}.
    
    \item GoPro. GoPro Awards: Great Barrier Reef with Fusion Overcapture in 4K. 2018. (YouTube): \\
    \url{https://youtu.be/OAmBkfn62dY}.
    
    \item GoPro. GoPro: Freediving with Tiger Sharks in 4K. 2017. (YouTube): 
    \url{https://youtu.be/Zy3kdMFvxUU}.
    \item TFIL. SCUBA DIVING WITH SHARKS!. 2017. (YouTube):
    \url{https://youtu.be/v8eSPf4RzTU}.
    
    \item Vins and Annette Singh. Stunning salt Water Fishes in a Marine Aquarium. 2019. (YouTube): \\
    \url{https://youtu.be/CWzXL6a4KGM}.
    
    %% Wrecks
    \item Akouris. H.M.Submarine Perseus. 2014. (YouTube):
    \url{https://youtu.be/4-oP0sX723k}.
    
    \item Gung Ho Vids. U.S. Navy Divers View An Underwater Wreck. 2014. (YouTube): \\
    \url{https://youtu.be/1qfRQRUMnXY}.
    
    \item Martcerv. Truk lagoon deep wrecks, GoPro black with SRP tray and lights. 2013. (YouTube): \\
    \url{https://youtu.be/0uD-nCN03s8}.
    
    \item Dmireiy. Shipwreck Diving, Nassau Bahamas. 2012. (YouTube):
    \url{https://youtu.be/CIQI3isddbE}.
    
    \item Frank Lame. diving WWII Wrecks around Palau. 2010. (YouTube):
    \url{https://youtu.be/vcI63XQsNlI}.
    
    \item Stevanurk. Wreck Dives Malta. 2014. (YouTube): \\
    \url{https://youtu.be/IZFuOIwEBH8}.
    
    \item Stevanurk. Diving Malta, Gozo and Comino 2015 Wrecks Caves. 2015. (YouTube): \\
    \url{https://youtu.be/NrDDjnij7sA}.
    
    \item Octavio velazquez lozano. SHIPWRECK Scuba Diving BAHAMAS. 2017. (YouTube):\\
    \url{https://youtu.be/4ovFPCEw4Qk}.
    
    \item Drew Kaplan. SCUBA Diving The Sunken Ancient Roman City Of Baiae, the Underwater Pompeii. 2018. (YouTube): \\
    \url{https://youtu.be/8RmJ3jzrwH8}.
    
    \item Octavio velazquez lozano. LIBERTY SHIPWRECK scuba dive destin florida. 2017. (YouTube): \\
    \url{https://youtu.be/DHuHZdVWONk}.
    
    \item Blue Robotics. BlueROV2 Dive: Hawaiian Open Water. 2016. (YouTube): \\
    \url{https://youtu.be/574jPVEk7mo}.
    
    \item JerryRigEverything. Exploring a Plane Wreck - UNDER WATER!. 2018. (YouTube): \\
    \url{https://youtu.be/0-sZVJbUzqo}.
    
    %% robots
    \item Rovrobotsubmariner. Home-built Underwater Robot ROV in Action!. 2010. (YouTube): \\
    \url{https://youtu.be/khLEyyf3Ci8}.
    
    \item Oded Ezra. Eca-Robotics H800 ROV. 2016. (YouTube): \\
    \url{https://youtu.be/Yafq9c7cqgE}.
    
    \item Geneinno Tech. Titan Diving Drone. 2019. (YouTube): \\
    \url{https://youtu.be/h7Bn4MxkFxs}.
    
    \item Scubo. Scubo - Agile Multifunctional Underwater Robot - ETH Zurich. 2016. (YouTube): \\
    \url{https://youtu.be/-g2O8e1j3fw}.
    
    \item Learning with Shedd. Student-built Underwater Robot at Shedd ROV Club Event. 2017. (YouTube): \\
    \url{https://youtu.be/y3dn8snT8os}.
    
    \item HMU-CSRL. SQUIDBOT sea trials. 2015. (YouTube): \\
    \url{https://youtu.be/0iDBF23gI6I}. 
    
    \item MobileRobots. Aqua2 Underwater Robot Navigates in a Coral Reef - Barbados. 2012. (YouTube): \\
    \url{https://youtu.be/jC-AmPfInwU}.
    
    \item Daniela Rus. underwater robot. 2015. (YouTube): \\
    \url{https://youtu.be/neLu0ZGuXPM}.
    
    \item JohnFardoulis. Sirius - Underwater Robot, Mapping. 2014. (YouTube): 
    \url{https://youtu.be/fXxVcucOPrs}.

\end{enumerate}

}

\end{document}